%% file: main.tex
\documentclass[letterpaper, 10 pt, journal, twoside]{IEEEtran}
\ifCLASSINFOpdf
\else
\fi
\usepackage{color}
\usepackage{transparent}
\usepackage{import}
\usepackage{multirow}
\usepackage{balance}
\usepackage{graphicx,xcolor}
\usepackage{psfrag}
\usepackage{amsmath,amsfonts} 

\usepackage{amssymb}  
\usepackage{makecell}
\usepackage{hyperref}

\usepackage{booktabs,tabularx,caption,ragged2e}
\usepackage{cite}
\usepackage{times,xspace}
\usepackage[font={footnotesize}]{caption}
\usepackage{makecell}
\usepackage{algorithm} 
\usepackage{algpseudocode} 
\usepackage{nicefrac}
\usepackage{romannum}
\usepackage{subcaption}
\usepackage{siunitx}
\usepackage{url}
\usepackage{array,multirow}
\usepackage[bottom]{footmisc}

\usepackage[makeroom]{cancel}

\usepackage{mathtools}

\usepackage{amssymb}  
\usepackage{makecell}
\usepackage{wrapfig}
\usepackage[frozencache,cachedir=minted-cache]{minted}
\setminted{frame=lines, breaklines, breaksymbol={\,}, fontsize=\footnotesize, escapeinside=||, autogobble=true}

\usepackage{array}
\usepackage{stmaryrd}
\usepackage{accents}
\usepackage{ntheorem}
\theoremstyle{break}
\theoremheaderfont{\normalfont\bfseries}
\theoremseparator{:}

\RequirePackage[                    
strict=true,                    
style=german                    
]{csquotes}

\definecolor{RPTHred}{RGB}{227, 27, 35}
\definecolor{TUMblue}{RGB}{0, 101, 189}
\definecolor{codegreen}{rgb}{0,0.6,0}
\definecolor{openaigreen}{RGB}{115,169,156}
\definecolor{openaiblue}{RGB}{115,140,159}
\definecolor{openaired}{RGB}{159,115,118}
\definecolor{codeplanner}{RGB}{100,160,200}
\definecolor{codegray}{RGB}{221,226,230}
\definecolor{codepurple}{rgb}{0.5,0,0.9}
\definecolor{backcolour}{rgb}{0.95,0.95,0.95}
\definecolor{initialplanner}{RGB}{253,243,224}
\definecolor{firstiteration}{RGB}{249,207,128}
\definecolor{seconditeration}{RGB}{226, 168, 79}
\definecolor{thirditeration}{RGB}{202,128,30}
\definecolor{fourceiteration}{RGB}{182, 115, 27}
\definecolor{codegreen2}{RGB}{81,140,66}
\definecolor{TUMbluelight}{RGB}{194, 215, 239}
\usepackage{listings} 
\usepackage{xcolor} 
\usepackage{fix-cm} 
\usepackage{tikz} 
\newcommand\copyrighttext{%
	\footnotesize \copyright 2024 IEEE. Personal use of this material is permitted. Permission from IEEE must be obtained for all other uses, in any current or future media, including reprinting/republishing this material for advertising or promotional purposes, creating new collective works, for resale or redistribution to servers or lists, or reuse of any copyrighted component of this work in other works.}
\newcommand\copyrightnotice{%
	\begin{tikzpicture}[remember picture,overlay]
		\node[anchor=south,yshift=5pt] at (current page.south) {\fbox{\parbox{\dimexpr\textwidth-\fboxsep-\fboxrule\relax}{\copyrighttext}}};
	\end{tikzpicture}%
}
\lstdefinestyle{mystyle}{
	backgroundcolor=\color{TUMbluelight},   
	commentstyle=\color{openaigreen},
	keywordstyle=\color{magenta},
	numberstyle=\tiny\color{openaigreen},
	stringstyle=\color{codepurple},
	basicstyle=\fontsize{7.5}{8}\selectfont\ttfamily\ttfamily,
	breakatwhitespace=false,         
	breaklines=true,
	breakindent=0pt,
	captionpos=t,                    
	keepspaces=true,                 
	numbers=none,                    
	numbersep=5pt,                  
	showspaces=false,                
	showstringspaces=false,
	showtabs=false,                  
	tabsize=2,
	framesep=5pt, 
	xleftmargin=15pt, 
	xrightmargin=15pt, 
	framexleftmargin=10pt, 
	framexrightmargin=10pt, 
}

\usepackage{bm}
\usepackage{colortbl}
\usepackage{pifont}
\usepackage{soul}
\usepackage{tikz}
\usetikzlibrary{calc}

\usepackage{hyperref}
\usepackage{xcolor}

\definecolor{RPTHred}{RGB}{227, 27, 35}
\definecolor{TUMblue}{RGB}{0, 101, 189}
\usepackage{romannum}

\usepackage{pifont}

\setquotestyle{english}

\begin{document}
%
\title{DrPlanner:
	Diagnosis and Repair of \\Motion Planners for Automated Vehicles Using Large Language Models}
%
%
%


\author{Yuanfei~Lin$^{1}$,~\IEEEmembership{Student Member,~IEEE,}  Chenran Li$^{2}$, Mingyu Ding$^{2}$,~\IEEEmembership{Student Member,~IEEE,}\\ Masayoshi Tomizuka$^{2}$,~\IEEEmembership{Life Fellow,~IEEE}, Wei Zhan$^{2}$,~\IEEEmembership{Member,~IEEE,} and Matthias Althoff$^{1}$,~\IEEEmembership{Member,~IEEE}%

\thanks{Manuscript received: March 12, 2024; Revised June 08, 2024; Accepted July 24, 2024. 
	This paper was recommended for publication by Editor  H. Kurniawati upon evaluation of the Associate Editor and Reviewers' comments. 
The authors gratefully acknowledge partial financial support by the German Federal Ministry for Digital and Transport (BMDV) for the project KoSi,
by the Deutsche Forschungsgemeinschaft (DFG, German Research Foundation) for the project SFB 1608,  under Grant 501798263,
 and by the Berkeley DeepDrive. The work was developed during Y. Lin’s visit to the University of California, Berkeley. \textit{(Corresponding
	author: Wei Zhan.)}} 

\thanks{$^1$ Y. Lin and M. Althoff are with the School of Computation, Information and Technology, Technical University of Munich, 85748 Garching, Germany. (e-mail: {\tt\footnotesize yuanfei.lin@tum.de}; {\tt\footnotesize{althoff@tum.de})}.}

\thanks{$^2$ C. Li, M. Ding, M. Tomizuka, and W. Zhan are with the Department
	of Mechanical Engineering, University of California, Berkeley, CA 94720, USA.
	(e-mail: {\tt\footnotesize chenran\_li@berkeley.edu}; {\tt\footnotesize myding@berkeley.edu}; {\tt\footnotesize tomizuka@berkeley.edu}; {\tt\footnotesize{wzhan@berkeley.edu})}.}

\thanks{Digital Object Identifier (DOI): see top of this page.}
}
%
%

\markboth{IEEE Robotics and Automation Letters. Preprint Version. Accepted July, 2024}
{Lin \MakeLowercase{\textit{et al.}}: DrPlanner:
	{D}iagnosis and {R}epair of 
	Motion {Planner}s {for Automated Vehicles} Using Large Language Models} 

%



\maketitle

\begin{abstract}
		Motion planners are essential for the safe operation of automated vehicles across various scenarios. However, no motion planning algorithm has achieved perfection in the literature, and improving its performance is often time-consuming and labor-intensive.
To tackle the aforementioned issues, we present ${\mathtt{DrPlanner}}$, the first framework designed to automatically \underline{d}iagnose and \underline{r}epair motion \underline{planner}s using large language models. Initially, we generate a structured description of the planner and its planned trajectories from both natural and programming languages.
Leveraging the profound capabilities of large language models, our framework returns repaired planners with detailed diagnostic descriptions.
Furthermore, our framework advances iteratively with continuous feedback from the evaluation of the repaired outcomes. 
Our approach is validated using both search- and sampling-based motion planners for automated vehicles; experimental results highlight the need for demonstrations in the prompt and show the ability of our framework to effectively identify and rectify elusive issues. 
\end{abstract}


	\input{content/introduction_diag}

\input{content/preliminaries}
	\input{content/diagnosis}
\input{content/evaluation}

	\input{content/conclusion}
%
\IEEEpeerreviewmaketitle

\section*{Acknowledgment}
The authors kindly thank Sebastian Illing for implementing the experiments for the sampling-based planner.

\ifCLASSOPTIONcaptionsoff
  \newpage
\fi

\bibliographystyle{IEEEtran}
	\IEEEtriggeratref{45}
{\bibliography{diagnosis.bib}}
\end{document}

%% file: content/introduction_diag.tex
\section{Introduction}\label{sec:introduction}
\begin{figure}[!t]%
	\centering
	\vspace{1mm}
	\def\svgwidth{1\columnwidth}\footnotesize
	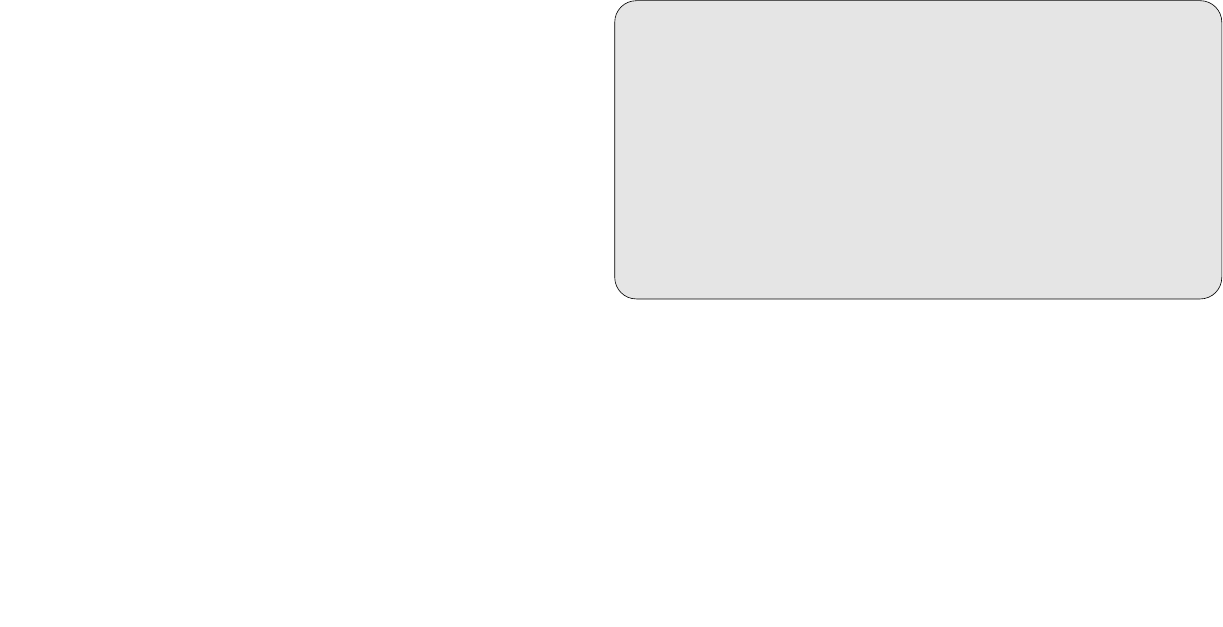
	 \caption{An example usage of  $\mathtt{DrPlanner}$: In a critical scenario, our imperfect motion planner plans a trajectory. The description of the trajectory and the planner is then fed into $\mathtt{DrPlanner}$. By harnessing the strengths of LLMs, 
	 	we adeptly diagnose and repair the deficiencies within the planner. }\label{fig:schema_general}
	\vspace{-5mm}
\end{figure}

\IEEEPARstart{M}{otion} planners for automated vehicles are responsible for computing safe, physically feasible, and comfortable motions~\cite{Paden2016}. 
 A major challenge is the excessive manual effort required to tune motion planners, which entails diagnosing the planner based on a variety of critical test scenarios and evaluation metrics. \copyrightnotice
To address this, we establish a framework that leverages the remarkable emergent abilities of large language models (LLMs)  
 \cite{brown2020language, ouyang2022training, openai2023gpt4} to automatically provide and apply diagnostic solutions for a motion planner {of automated vehicles}, as illustrated in  Fig.~\ref{fig:schema_general}.      

\subsection{Related Work}\label{sec:li_ov}

Although many motion planning algorithms can tackle a diverse range of tasks, they often face issues related to probabilistic completeness, computational complexity, or real-time constraints in finding the optimal solution \cite{zucker2013chomp, Paden2016, gu2016runtime, Aradi2020, liu2022benchmarking}.  Besides, guaranteeing safety, rule compliance, and social compatibility of motion planners remains a challenge \cite{Krasowski2020a, wang2021socially, YuanfeiLinMPR, mehdipour2023formal}. To provide an overview of how one can improve and repair such planners, we first survey methods from automated software repair, followed by summarizing contributions based on LLMs.


\subsubsection{\it Automated Software Repair}\label{subsec:apr}
With the increasing complexity and size of software, automatic debugging and repair techniques have been developed to reduce the extensive manual effort required to fix faults and to improve quality~\cite{gazzola2018automatic}. For instance, human-designed templates are used to repair certain types of bugs in code \cite{weimer2009automatically, kim2013automatic, liu2018mining, liu2019tbar, koyuncu2020fixminer}, but their effectiveness is often limited to hard-coded patterns. To overcome these limitations, deep-learning-based approaches utilize neural machine translation \cite{bahdanau2014neural} to learn from existing patches, treating the repaired code as a translation of the buggy one \cite{lutellier2020coconut, zhu2021syntax, jiang2021cure, ye2022neural}.
However, the performance of these approaches is limited by the quality and quantity of the training data as well as its representation format~\cite{xia2022less}. As LLMs have shown emergent abilities in solving programming tasks   \cite{feng2020codebert, chen2021evaluating, austin2021program, fried2022incoder, liang2023code}, they are applied for generating program patches \cite{kolak2022patch, prenner2022can, xia2022practical}, self-debugging \cite{shinn2023reflexion, chen2023teaching}, and cleaning code~\cite{jain2023llm}. 
Unlike simply maintaining functional equivalence, we aim to both rectify imperfections and boost the performance of the planning algorithms. 
{Although the aspect of linking text with code aligns with \cite{liang2023code} and the focus on performance improvement with \cite{madaan2023learning}, our work uniquely addresses the challenges posed by the larger and more intricate codebases of motion planners.}
Another branch of work focuses on  repairing the outcome of given software \cite{YuanfeiLin2021, YuanfeiLin2022a,  maierhofer2023map} or addressing specified diagnostic criteria~\cite{pacheck2023physically}.
\subsubsection {\it Language Models for Motion Planning}\label{subsec:llm}
With their indispensable role of common sense reasoning and generalization \cite{kojima2022large, yao2023react, kiciman2023causal}, LLMs have been applied in motion planning for autonomous driving to make high-level decisions \cite{sha2023languagempc, wen2023dilu, wang2023drivemlm, sima2023drivelm,  cui2024drive}, generate driving trajectories \cite{mao2023gpt, mao2023language} or provide control signals directly \cite{xu2023drivegpt4, chen2023driving, shao2023lmdrive}.
However, the refinement of motion planners themselves is still driven by the nuanced intuition of humans and by real traffic data. In this work, LLMs serve to bridge this gap by emulating human-like problem-solving strategies, offering strategic guidance in analyzing complex motion planners.

\subsection{Contributions}\label{sec:contri}
In this work, we introduce ${\mathtt{DrPlanner}}$, the first framework to autonomously \underline{d}iagnose and \underline{r}epair motion \underline{planner}s {for automated vehicles}, harnessing the power of LLMs that improve as they scale with additional data and model complexity.
In particular, our contributions are:

\begin{enumerate}
	\item establishing a structured and modular description for motion planners across both natural and programming language modalities  to exploit the capabilities of LLMs for diagnosis and repair;

	\item  leveraging the in-context learning capabilities of LLMs by providing demonstrations to the model at the point where it infers diagnostic results; and
		\item  enhancing the understanding of underlying improvement mechanisms by generating continuous feedback in a closed-loop manner. 
\end{enumerate}
The remainder of this work is structured as follows:
Sec.~\ref{sec:pre} lists necessary preliminaries. The proposed framework for diagnosing and repairing motion planners is described in  Sec.~\ref{sec:dr}. We demonstrate the benefits of our approach in Sec.~\ref{sec:eva} and conclude the paper in Sec.~\ref{sec:conc}.
%
%

%% file: figures/framework_diag_4.pdf_tex
\begingroup%
  \makeatletter%
  \providecommand\color[2][]{%
    \errmessage{(Inkscape) Color is used for the text in Inkscape, but the package 'color.sty' is not loaded}%
    \renewcommand\color[2][]{}%
  }%
  \providecommand\transparent[1]{%
    \errmessage{(Inkscape) Transparency is used (non-zero) for the text in Inkscape, but the package 'transparent.sty' is not loaded}%
    \renewcommand\transparent[1]{}%
  }%
  \providecommand\rotatebox[2]{#2}%
  \newcommand*\fsize{\dimexpr\f@size pt\relax}%
  \newcommand*\lineheight[1]{\fontsize{\fsize}{#1\fsize}\selectfont}%
  \ifx\svgwidth\undefined%
    \setlength{\unitlength}{586.62087268bp}%
    \ifx\svgscale\undefined%
      \relax%
    \else%
      \setlength{\unitlength}{\unitlength * \real{\svgscale}}%
    \fi%
  \else%
    \setlength{\unitlength}{\svgwidth}%
  \fi%
  \global\let\svgwidth\undefined%
  \global\let\svgscale\undefined%
  \makeatother%
  \begin{picture}(1,0.51237404)%
    \lineheight{1}%
    \setlength\tabcolsep{0pt}%
    \put(0,0){\includegraphics[width=\unitlength,page=1]{framework_diag_4.pdf}}%
    \put(0.56766595,0.48026591){\color[rgb]{0,0,0}\makebox(0,0)[lt]{\lineheight{1.25}\smash{\begin{tabular}[t]{l}\textbf{Imperfect Motion Planner}\end{tabular}}}}%
    \put(0,0){\includegraphics[width=\unitlength,page=2]{framework_diag_4.pdf}}%
    \put(0.65237872,0.19592301){\color[rgb]{1,1,1}\makebox(0,0)[lt]{\lineheight{1.25}\smash{\begin{tabular}[t]{l}\textbf{LLM}\end{tabular}}}}%
    \put(0,0){\includegraphics[width=\unitlength,page=3]{framework_diag_4.pdf}}%
    \put(0.03949711,0.114033){\color[rgb]{0,0,0}\makebox(0,0)[lt]{\lineheight{1.25}\smash{\begin{tabular}[t]{l}\scriptsize{\textbf{Diagnosis:}\textit{ Cost calculation includes negative value}}\end{tabular}}}}%
    \put(0.03897496,0.08225304){\color[rgb]{0,0,0}\makebox(0,0)[lt]{\lineheight{1.25}\smash{\begin{tabular}[t]{l}\scriptsize{\textbf{Prescription:}\textit{ Ensure cost is non-negative by including condition that sets cost}}\end{tabular}}}}%
    \put(0,0){\includegraphics[width=\unitlength,page=4]{framework_diag_4.pdf}}%
    \put(0.04057582,0.01462936){\color[rgb]{0,0,0}\makebox(0,0)[lt]{\lineheight{1.25}\smash{\begin{tabular}[t]{l}\scriptsize{\textbf{Repaired planner:} $\mathtt{def\ heuristic\_function(...)/cost\_function(...)/...}$}\end{tabular}}}}%
    \put(0.20721328,0.05551998){\color[rgb]{0,0,0}\makebox(0,0)[lt]{\lineheight{1.25}\smash{\begin{tabular}[t]{l}\textit{\scriptsize{to zero if negative}\textit{}}\end{tabular}}}}%
    \put(0,0){\includegraphics[width=\unitlength,page=5]{framework_diag_4.pdf}}%
    \put(0.2236791,0.19732157){\color[rgb]{0,0,0}\makebox(0,0)[lt]{\lineheight{1.25}\smash{\begin{tabular}[t]{l}\textbf{${\mathtt{DrPlanner}}$}\end{tabular}}}}%
    \put(0,0){\includegraphics[width=\unitlength,page=6]{framework_diag_4.pdf}}%
    \put(0.13085026,0.47974536){\color[rgb]{0,0,0}\makebox(0,0)[lt]{\lineheight{1.25}\smash{\begin{tabular}[t]{l}\textbf{Critical Scenario}\end{tabular}}}}%
    \put(0,0){\includegraphics[width=\unitlength,page=7]{framework_diag_4.pdf}}%
    \put(0.13956209,0.3216059){\color[rgb]{0,0,0}\makebox(0,0)[lt]{\lineheight{1.25}\smash{\begin{tabular}[t]{l}\scriptsize{initial state}\end{tabular}}}}%
    \put(0.3085703,0.36828488){\color[rgb]{0,0,0}\makebox(0,0)[lt]{\lineheight{1.25}\smash{\begin{tabular}[t]{l}\scriptsize{goal region}\end{tabular}}}}%
    \put(0,0){\includegraphics[width=\unitlength,page=8]{framework_diag_4.pdf}}%
    \put(0.17545689,0.36829395){\color[rgb]{0,0,0}\makebox(0,0)[lt]{\lineheight{1.25}\smash{\begin{tabular}[t]{l}\scriptsize{obstacles}\end{tabular}}}}%
    \put(0,0){\includegraphics[width=\unitlength,page=9]{framework_diag_4.pdf}}%
    \put(0.57637572,0.43625224){\color[rgb]{0,0,0}\makebox(0,0)[lt]{\lineheight{1.25}\smash{\begin{tabular}[t]{l}\scriptsize{search-based}\end{tabular}}}}%
    \put(0,0){\includegraphics[width=\unitlength,page=10]{framework_diag_4.pdf}}%
    \put(0.71884613,0.30225404){\color[rgb]{0,0,0}\makebox(0,0)[lt]{\lineheight{1.25}\smash{\begin{tabular}[t]{l}\scriptsize{sampling-based}\end{tabular}}}}%
    \put(0,0){\includegraphics[width=\unitlength,page=11]{framework_diag_4.pdf}}%
    \put(0.92556775,0.30801807){\color[rgb]{0,0,0}\makebox(0,0)[lt]{\lineheight{1.25}\smash{\begin{tabular}[t]{l}\scriptsize{...}\end{tabular}}}}%
  \end{picture}%
\endgroup%

%% file: content/preliminaries.tex
\section{Preliminaries}\label{sec:pre}
\subsection{Motion Planning {for Automated Vehicles}}\label{subsec:mp}
\begin{figure}[!t]%
	\centering
	\vspace{0.5mm}
	\def\svgwidth{0.85\columnwidth}\footnotesize
	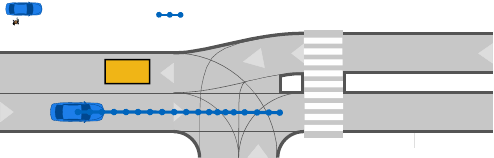
	\caption{Exemplary motion planning problem, where the ego vehicle needs to travel from its initial state to reach the goal region safely and efficiently.}\label{fig:cr_scenario}
	\vspace{-6mm}
\end{figure}

We refer to the vehicle for which trajectories are planned as the \textit{ego vehicle}. 
As illustrated in Fig.~\ref{fig:cr_scenario}, motion planning algorithms are tasked with ensuring that the ego vehicle travels from an initial state to a goal region within a specified time~\cite{Althoff2017a}. The motion planner typically minimizes a given objective function $J(\chi)$, e.g., by penalizing the travel time or passenger discomfort \cite[Sec. IV]{Paden2016}. Simultaneously, the solution, denoted by $\chi$, must satisfy common and safety-relevant requirements, such as being drivable, collision-free, and rule-compliant \cite{PekIV20, YuanfeiLin2022a}. 
Subsequently, we denote a motion planner by $\mathtt{M}$ and a motion planning problem by  $\mathtt{{P}}$. 

\subsection{Prompt Engineering for LLMs}\label{subsec:prompt_eng}
The technique of using a textual string $\ell$ to instruct LLMs 
is referred to as \textit{prompting}~\cite[Sec.~4]{liu2023pre}. This approach enables LLMs to be pretrained on a massive amount of data ~\cite[Sec.~3]{liu2023pre} and subsequently adapt to new use cases with few or no labeled data.
To enhance the in-context learning capabilities, the prompt may include a few human-annotated examples of the task, known as \textit{few-shot prompting}~\cite{brown2020language}, or utilize chain-of-thought reasoning \cite{wei2022chain, kojima2022large}. 
We divide the input prompt $\ell$ into two components: the system prompt $\ell_{\text{system}}$, which outlines the task for the LLMs, and the user prompt $\ell_{\text{user}}$, providing context for the diagnostic task. 
{The labels, manual inputs, and automatically generated content within the prompt are marked with angle brackets, square brackets, and curly brackets, respectively.}
The output consists of both a list of diagnosis-prescription pairs and patched programs, collectively denoted by $\boldsymbol{\ell}_{\text{dp}}$ and $\boldsymbol{p}_{\text{p}}$. 
It is important to consider that LLMs have a limit on the number of tokens they can process \cite{vaswani2017attention}, which {imposes a maximum length on the prompt and prevents us from including extensive code within a single prompt}.

\begin{figure*}[!t]%
	\centering
	\vspace{1.5mm}
	\def\svgwidth{1.985\columnwidth}\footnotesize
	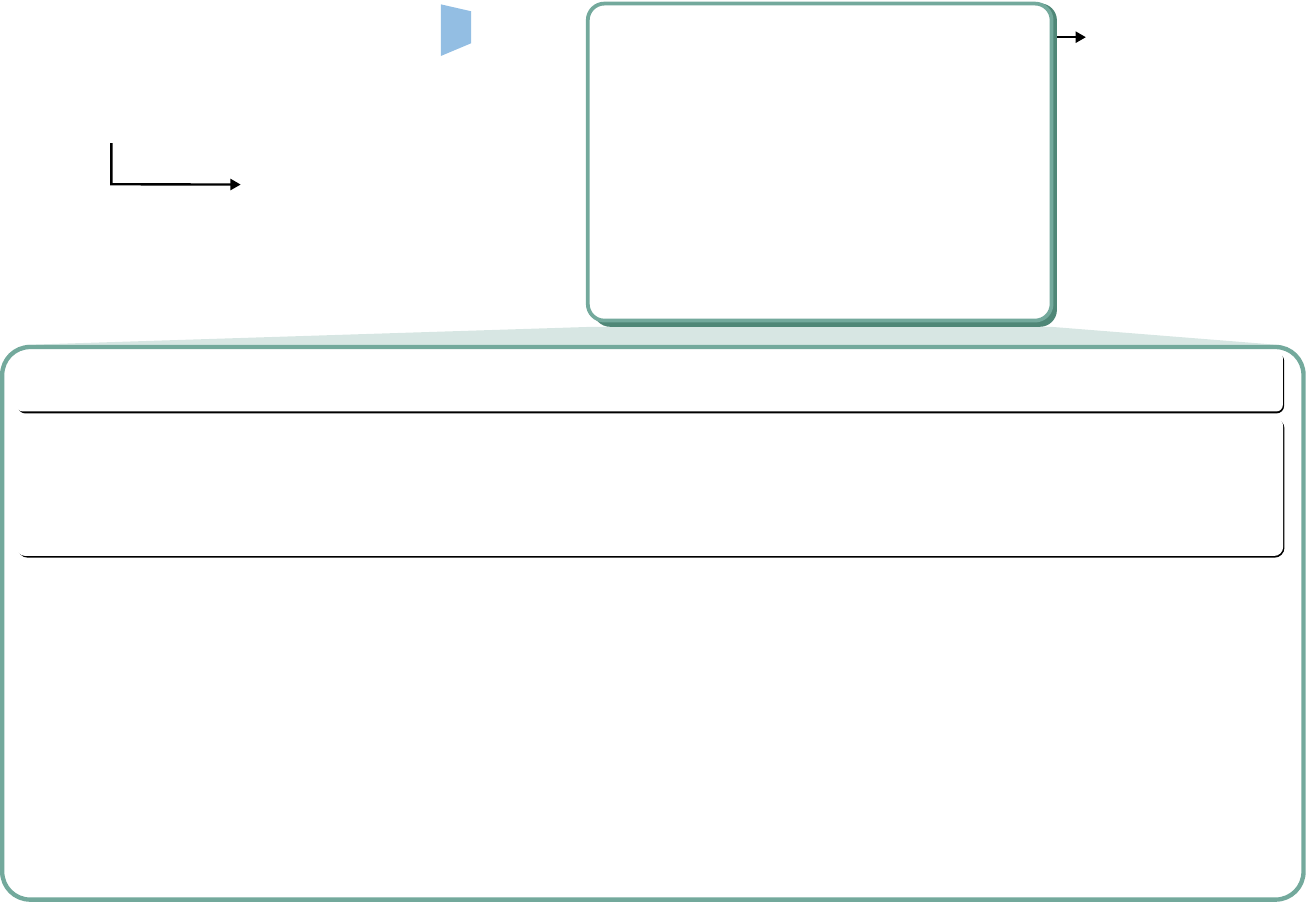
	\caption{Overview of the $\mathtt{DrPlanner}$ framework. The process starts with obtaining a planned trajectory for the planning problem with the given motion planner. Then, the planned trajectory is evaluated by the objective function. Afterwards, the description for the planner is generated and used to prompt an off-the-shelf LLM to generate the diagnoses and prescriptions for the planner, along with the patched programs. After applying the patches, the evaluation of the updated planner is incorporated back into the prompt as feedback to continuously enhance the diagnostic performance (marked by dashed arrows).}	\label{fig:framework}
	\vspace{-1.5mm}
\end{figure*}%

%% file: figures/cr_scenario.pdf_tex
\begingroup%
  \makeatletter%
  \providecommand\color[2][]{%
    \errmessage{(Inkscape) Color is used for the text in Inkscape, but the package 'color.sty' is not loaded}%
    \renewcommand\color[2][]{}%
  }%
  \providecommand\transparent[1]{%
    \errmessage{(Inkscape) Transparency is used (non-zero) for the text in Inkscape, but the package 'transparent.sty' is not loaded}%
    \renewcommand\transparent[1]{}%
  }%
  \providecommand\rotatebox[2]{#2}%
  \newcommand*\fsize{\dimexpr\f@size pt\relax}%
  \newcommand*\lineheight[1]{\fontsize{\fsize}{#1\fsize}\selectfont}%
  \ifx\svgwidth\undefined%
    \setlength{\unitlength}{236.74215118bp}%
    \ifx\svgscale\undefined%
      \relax%
    \else%
      \setlength{\unitlength}{\unitlength * \real{\svgscale}}%
    \fi%
  \else%
    \setlength{\unitlength}{\svgwidth}%
  \fi%
  \global\let\svgwidth\undefined%
  \global\let\svgscale\undefined%
  \makeatother%
  \begin{picture}(1,0.31989481)%
    \lineheight{1}%
    \setlength\tabcolsep{0pt}%
    \put(0,0){\includegraphics[width=\unitlength,page=1]{cr_scenario.pdf}}%
    \put(0.09455003,0.28370936){\color[rgb]{0,0,0}\makebox(0,0)[lt]{\lineheight{1.25}\smash{\begin{tabular}[t]{l}obstacles\end{tabular}}}}%
    \put(0.6831992,0.30036192){\color[rgb]{0,0,0}\makebox(0,0)[lt]{\lineheight{1.25}\smash{\begin{tabular}[t]{l}initial\end{tabular}}}}%
    \put(0.68319335,0.26554459){\color[rgb]{0,0,0}\makebox(0,0)[lt]{\lineheight{1.25}\smash{\begin{tabular}[t]{l}state\end{tabular}}}}%
    \put(0.90309565,0.30282419){\color[rgb]{0,0,0}\makebox(0,0)[lt]{\lineheight{1.25}\smash{\begin{tabular}[t]{l}goal\end{tabular}}}}%
    \put(0.90345192,0.26638629){\color[rgb]{0,0,0}\makebox(0,0)[lt]{\lineheight{1.25}\smash{\begin{tabular}[t]{l}region\end{tabular}}}}%
    \put(0,0){\includegraphics[width=\unitlength,page=2]{cr_scenario.pdf}}%
    \put(0.38351131,0.30121496){\color[rgb]{0,0,0}\makebox(0,0)[lt]{\lineheight{1.25}\smash{\begin{tabular}[t]{l}future\end{tabular}}}}%
    \put(0.38386739,0.26477705){\color[rgb]{0,0,0}\makebox(0,0)[lt]{\lineheight{1.25}\smash{\begin{tabular}[t]{l}movement\end{tabular}}}}%
    \put(0,0){\includegraphics[width=\unitlength,page=3]{cr_scenario.pdf}}%
  \end{picture}%
\endgroup%

%% file: figures/pipeline8.pdf_tex
\begingroup%
  \makeatletter%
  \providecommand\color[2][]{%
    \errmessage{(Inkscape) Color is used for the text in Inkscape, but the package 'color.sty' is not loaded}%
    \renewcommand\color[2][]{}%
  }%
  \providecommand\transparent[1]{%
    \errmessage{(Inkscape) Transparency is used (non-zero) for the text in Inkscape, but the package 'transparent.sty' is not loaded}%
    \renewcommand\transparent[1]{}%
  }%
  \providecommand\rotatebox[2]{#2}%
  \newcommand*\fsize{\dimexpr\f@size pt\relax}%
  \newcommand*\lineheight[1]{\fontsize{\fsize}{#1\fsize}\selectfont}%
  \ifx\svgwidth\undefined%
    \setlength{\unitlength}{626.74393782bp}%
    \ifx\svgscale\undefined%
      \relax%
    \else%
      \setlength{\unitlength}{\unitlength * \real{\svgscale}}%
    \fi%
  \else%
    \setlength{\unitlength}{\svgwidth}%
  \fi%
  \global\let\svgwidth\undefined%
  \global\let\svgscale\undefined%
  \makeatother%
  \begin{picture}(1,0.69066391)%
    \lineheight{1}%
    \setlength\tabcolsep{0pt}%
    \put(0,0){\includegraphics[width=\unitlength,page=1]{pipeline8.pdf}}%
    \put(0.75172548,0.64896476){\color[rgb]{0.41568627,0.45882353,0.49411765}\makebox(0,0)[lt]{\lineheight{1.25}\smash{\begin{tabular}[t]{l}$\ell_{\text{system}}$\end{tabular}}}}%
    \put(0.75946565,0.62224863){\color[rgb]{0.41568627,0.45882353,0.49411765}\makebox(0,0)[lt]{\lineheight{1.25}\smash{\begin{tabular}[t]{l}$\ell_{\text{user}}$\end{tabular}}}}%
    \put(0.46044145,0.54298418){\color[rgb]{0.92156863,0.44313725,0.23529412}\makebox(0,0)[lt]{\lineheight{1.25}\smash{\begin{tabular}[t]{l}\textbf{\texttt{<planned trajectory>} (cf. Sec.~\ref{subsec:des_traj})}\end{tabular}}}}%
    \put(0.46193673,0.50172711){\color[rgb]{0.45098039,0.6627451,0.61176471}\makebox(0,0)[lt]{\lineheight{1.25}\smash{\begin{tabular}[t]{l}\textbf{\texttt{<few-shots>} (cf. Sec.~\ref{subsec:few-shots})}\end{tabular}}}}%
    \put(0.46151465,0.46081935){\color[rgb]{0.41568627,0.45882353,0.49411765}\makebox(0,0)[lt]{\lineheight{1.25}\smash{\begin{tabular}[t]{l}\textbf{\texttt{<feedback>} (cf. Sec.~\ref{subsec:reprompt})}\end{tabular}}}}%
    \put(0.46019241,0.64889115){\color[rgb]{0,0,0}\makebox(0,0)[lt]{\lineheight{1.25}\smash{\begin{tabular}[t]{l}\texttt{<system>}\end{tabular}}}}%
    \put(0.4598727,0.62094695){\color[rgb]{0,0,0}\makebox(0,0)[lt]{\lineheight{1.25}\smash{\begin{tabular}[t]{l}\texttt{<instructions>} (cf. Sec.~\ref{subsec:dia_instruction})\end{tabular}}}}%
    \put(0,0){\includegraphics[width=\unitlength,page=2]{pipeline8.pdf}}%
    \put(0.90296337,0.66418368){\color[rgb]{1,1,1}\makebox(0,0)[lt]{\lineheight{1.25}\smash{\begin{tabular}[t]{l}\textbf{LLM}\end{tabular}}}}%
    \put(0,0){\includegraphics[width=\unitlength,page=3]{pipeline8.pdf}}%
    \put(0.9080151,0.66547766){\color[rgb]{1,1,1}\makebox(0,0)[lt]{\lineheight{1.25}\smash{\begin{tabular}[t]{l}\textbf{LLM}\end{tabular}}}}%
    \put(0,0){\includegraphics[width=\unitlength,page=4]{pipeline8.pdf}}%
    \put(0.83602968,0.60651897){\color[rgb]{0,0,0}\makebox(0,0)[lt]{\lineheight{1.25}\smash{\begin{tabular}[t]{l}\textbf{diagnosis}\end{tabular}}}}%
    \put(0.8362743,0.55303065){\color[rgb]{0,0,0}\makebox(0,0)[lt]{\lineheight{1.25}\smash{\begin{tabular}[t]{l}\textbf{prescription}\end{tabular}}}}%
    \put(0,0){\includegraphics[width=\unitlength,page=5]{pipeline8.pdf}}%
    \put(0.83706978,0.50078908){\color[rgb]{0,0,0}\makebox(0,0)[lt]{\lineheight{1.25}\smash{\begin{tabular}[t]{l}\textbf{patched program}\end{tabular}}}}%
    \put(0,0){\includegraphics[width=\unitlength,page=6]{pipeline8.pdf}}%
    \put(0.59876631,0.67259515){\color[rgb]{1,1,1}\makebox(0,0)[lt]{\lineheight{1.25}\smash{\begin{tabular}[t]{l}\textbf{Prompt}\end{tabular}}}}%
    \put(0,0){\includegraphics[width=\unitlength,page=7]{pipeline8.pdf}}%
    \put(0.33575434,0.46572765){\color[rgb]{1,1,1}\makebox(0,0)[lt]{\lineheight{1.25}\smash{\begin{tabular}[t]{l}Feedback\end{tabular}}}}%
    \put(0.33457192,0.44898748){\color[rgb]{1,1,1}\makebox(0,0)[lt]{\lineheight{1.25}\smash{\begin{tabular}[t]{l}Generator\end{tabular}}}}%
    \put(0.46028633,0.58426052){\color[rgb]{0.39215686,0.62745098,0.78431373}\makebox(0,0)[lt]{\lineheight{1.25}\smash{\begin{tabular}[t]{l}\textbf{\texttt{<motion planner>} (cf. Sec.~\ref{subsec:dia_planner})}\end{tabular}}}}%
    \put(0,0){\includegraphics[width=\unitlength,page=8]{pipeline8.pdf}}%
    \put(0.09256504,0.63138972){\color[rgb]{0,0,0}\makebox(0,0)[lt]{\lineheight{1.25}\smash{\begin{tabular}[t]{l}$\mathtt{P}$\end{tabular}}}}%
    \put(0,0){\includegraphics[width=\unitlength,page=9]{pipeline8.pdf}}%
    \put(0.06169937,0.60178085){\color[rgb]{0,0,0}\makebox(0,0)[lt]{\lineheight{1.25}\smash{\begin{tabular}[t]{l}\textbf{Motion}\end{tabular}}}}%
    \put(0.05944561,0.58504064){\color[rgb]{0,0,0}\makebox(0,0)[lt]{\lineheight{1.25}\smash{\begin{tabular}[t]{l}\textbf{Planner}\end{tabular}}}}%
    \put(0,0){\includegraphics[width=\unitlength,page=10]{pipeline8.pdf}}%
    \put(0.34037715,0.60077838){\color[rgb]{0,0,0}\makebox(0,0)[lt]{\lineheight{1.25}\smash{\begin{tabular}[t]{l}Planner\end{tabular}}}}%
    \put(0.33442404,0.58403804){\color[rgb]{0,0,0}\makebox(0,0)[lt]{\lineheight{1.25}\smash{\begin{tabular}[t]{l}Describer\end{tabular}}}}%
    \put(0,0){\includegraphics[width=\unitlength,page=11]{pipeline8.pdf}}%
    \put(0.15828595,0.59647888){\color[rgb]{0,0,0}\makebox(0,0)[lt]{\lineheight{1.25}\smash{\begin{tabular}[t]{l}key components\end{tabular}}}}%
    \put(0,0){\includegraphics[width=\unitlength,page=12]{pipeline8.pdf}}%
    \put(0.33303783,0.55601566){\color[rgb]{0,0,0}\makebox(0,0)[lt]{\lineheight{1.25}\smash{\begin{tabular}[t]{l}Trajectory\end{tabular}}}}%
    \put(0.33442404,0.53927541){\color[rgb]{0,0,0}\makebox(0,0)[lt]{\lineheight{1.25}\smash{\begin{tabular}[t]{l}Describer\end{tabular}}}}%
    \put(0,0){\includegraphics[width=\unitlength,page=13]{pipeline8.pdf}}%
    \put(0.3177399,0.51046445){\color[rgb]{0,0,0}\makebox(0,0)[lt]{\lineheight{1.25}\smash{\begin{tabular}[t]{l}Demonstration\end{tabular}}}}%
    \put(0.33442404,0.49372424){\color[rgb]{0,0,0}\makebox(0,0)[lt]{\lineheight{1.25}\smash{\begin{tabular}[t]{l}Describer\end{tabular}}}}%
    \put(0,0){\includegraphics[width=\unitlength,page=14]{pipeline8.pdf}}%
    \put(0.09222042,0.51247317){\color[rgb]{0,0,0}\makebox(0,0)[lt]{\lineheight{1.25}\smash{\begin{tabular}[t]{l}helper functions\end{tabular}}}}%
    \put(0.09256661,0.55693373){\color[rgb]{0,0,0}\makebox(0,0)[lt]{\lineheight{1.25}\smash{\begin{tabular}[t]{l}$\chi$\end{tabular}}}}%
    \put(0,0){\includegraphics[width=\unitlength,page=15]{pipeline8.pdf}}%
    \put(0.20253846,0.54392456){\color[rgb]{0,0,0}\makebox(0,0)[lt]{\lineheight{1.25}\smash{\begin{tabular}[t]{l}Evaluator\end{tabular}}}}%
    \put(0,0){\includegraphics[width=\unitlength,page=16]{pipeline8.pdf}}%
    \put(0.28366871,0.5534951){\color[rgb]{0,0,0}\makebox(0,0)[lt]{\lineheight{1.25}\smash{\begin{tabular}[t]{l}$J$\end{tabular}}}}%
    \put(0,0){\includegraphics[width=\unitlength,page=17]{pipeline8.pdf}}%
    \put(0.26367903,0.62034623){\color[rgb]{0,0,0}\makebox(0,0)[lt]{\lineheight{1.25}\smash{\begin{tabular}[t]{l}$J^*$\end{tabular}}}}%
    \put(0,0){\includegraphics[width=\unitlength,page=18]{pipeline8.pdf}}%
    \put(0.14670529,0.66365582){\color[rgb]{0,0,0}\makebox(0,0)[lt]{\lineheight{1.25}\smash{\begin{tabular}[t]{l}Leaderboard\end{tabular}}}}%
    \put(0,0){\includegraphics[width=\unitlength,page=19]{pipeline8.pdf}}%
    \put(0.26394404,0.67114415){\color[rgb]{0,0,0}\makebox(0,0)[lt]{\lineheight{1.25}\smash{\begin{tabular}[t]{l}Objective\end{tabular}}}}%
    \put(0.02192356,0.39465732){\color[rgb]{0,0,0}\makebox(0,0)[lt]{\lineheight{1.25}\smash{\begin{tabular}[t]{l}\makecell[l]{\texttt{<system>}: You are an expert in diagnosing motion planners for automated vehicles. Your task is to identify diagnoses and recommend \\prescriptions for the motion planner, with the objective of enhancing its performance.} \end{tabular}}}}%
    \put(0.02199219,0.31533996){\color[rgb]{0,0,0}\makebox(0,0)[lt]{\lineheight{1.25}\smash{\begin{tabular}[t]{l}\makecell[l]{\texttt{<instructions>}: Before you start, it is important to understand and adhere to the instructions: \\- Ensure that the improved code is free from errors and that all modifications positively impact the final outcome.\\- The diagnosis should concisely pinpoint each issue, using only a few words for clarity and brevity. For each prescription, provide a detailed,\\ step-by-step action plan. \textbf{[other instructions]}\\- Adhere strictly to all specified instructions. In case of a contradiction with your knowledge, offer a thorough explanation.} \end{tabular}}}}%
    \put(0,0){\includegraphics[width=\unitlength,page=20]{pipeline8.pdf}}%
    \put(0.0242144,0.22650599){\color[rgb]{0.39215686,0.62745098,0.78431373}\makebox(0,0)[lt]{\lineheight{1.25}\smash{\begin{tabular}[t]{l}\makecell[l]{\textbf{\texttt{<motion planner>}}: The \textbf{[planning algorithm]} is employed in trajectory planning to navigate the vehicle from an initial state to a \\designated goal region by \textbf{[principle]}. The key components of the planner to be diagnosed and repaired are: \textbf{[key component]} -  \\\textbf{[general description]} \textbf{\{code\} + \{detailed description\}}; ...} \end{tabular}}}}%
    \put(0,0){\includegraphics[width=\unitlength,page=21]{pipeline8.pdf}}%
    \put(0.02421424,0.16049742){\color[rgb]{0.92156863,0.44313725,0.23529412}\makebox(0,0)[lt]{\lineheight{1.25}\smash{\begin{tabular}[t]{l}\makecell[l]{\textbf{\texttt{<planned trajectory>}}: The goal is to adjust the total objective function of the planned trajectory to closely align with the desired \\value \textbf{[$\boldsymbol{J^*}$]}. The current total objective function is calculated to be \textbf{\{$\boldsymbol{J}$\}}, includes \textbf{\{component\}}, valued at \textbf{\{value\}} with a weight of\\\textbf{\{weight\}}; \textbf{\{component\}}, valued at ...} \end{tabular}}}}%
    \put(0,0){\includegraphics[width=\unitlength,page=22]{pipeline8.pdf}}%
    \put(0.02426035,0.09786447){\color[rgb]{0.45098039,0.6627451,0.61176471}\makebox(0,0)[lt]{\lineheight{1.25}\smash{\begin{tabular}[t]{l}\makecell[l]{\textbf{\texttt{<few-shots>}}: There are also some pre-defined helper functions that can be directly called in the \textbf{[key component]}: \textbf{[helper functions]} - \\\textbf{\{method definition + docstring\}}, \textbf{[examples]}; ... }\end{tabular}}}}%
    \put(0,0){\includegraphics[width=\unitlength,page=23]{pipeline8.pdf}}%
    \put(0.02426035,0.03866459){\color[rgb]{0.41568627,0.45882353,0.49411765}\makebox(0,0)[lt]{\lineheight{1.25}\smash{\begin{tabular}[t]{l}\makecell[l]{\textbf{\texttt{<feedback>}}: Diagnoses and prescriptions from the iteration \textbf{\{number of iteration\}}: \textbf{\{diagnoses and prescriptions\}}. After applying \\this diagnostic result, \textbf{\{error messages\}} / the updated total objective function is $\boldsymbol{\{J_{\text{\bf rep}}\}}$, which includes \textbf{\{details of the updated objective} \\\textbf{components\}}. The performance of the motion planner is getting \textbf{\{worse / better\}}. } \end{tabular}}}}%
    \put(0,0){\includegraphics[width=\unitlength,page=24]{pipeline8.pdf}}%
    \put(-3.67174872,-1.39217705){\color[rgb]{0,0,0}\makebox(0,0)[lt]{\begin{minipage}{0.66246596\unitlength}\raggedright \end{minipage}}}%
    \put(0.21079007,0.51424369){\color[rgb]{0.41568627,0.45882353,0.49411765}\makebox(0,0)[lt]{\lineheight{1.25}\smash{\begin{tabular}[t]{l}$J_{\text{rep}}$\end{tabular}}}}%
    \put(0,0){\includegraphics[width=\unitlength,page=25]{pipeline8.pdf}}%
    \put(0.26669832,0.65516084){\color[rgb]{0,0,0}\makebox(0,0)[lt]{\lineheight{1.25}\smash{\begin{tabular}[t]{l}Function\end{tabular}}}}%
    \put(0.05548228,0.6560396){\color[rgb]{0,0,0}\transparent{0.75}\makebox(0,0)[lt]{\lineheight{1.25}\smash{\begin{tabular}[t]{l}Scenario\end{tabular}}}}%
    \put(0.05756255,0.6706317){\color[rgb]{0,0,0}\transparent{0.75}\makebox(0,0)[lt]{\lineheight{1.25}\smash{\begin{tabular}[t]{l}Critical\end{tabular}}}}%
    \put(0,0){\includegraphics[width=\unitlength,page=26]{pipeline8.pdf}}%
    \put(0.26186131,0.47264276){\color[rgb]{0.41568627,0.45882353,0.49411765}\makebox(0,0)[lt]{\lineheight{1.25}\smash{\begin{tabular}[t]{l}$\boldsymbol{\ell}_{\text{dp}}$\end{tabular}}}}%
    \put(0.00179908,0.60031361){\color[rgb]{0.41568627,0.45882353,0.49411765}\makebox(0,0)[lt]{\lineheight{1.25}\smash{\begin{tabular}[t]{l} $\boldsymbol{p}_{\text{p}}$\end{tabular}}}}%
    \put(0,0){\includegraphics[width=\unitlength,page=27]{pipeline8.pdf}}%
    \put(0.12102141,0.59376764){\color[rgb]{0,0,0}\makebox(0,0)[lt]{\lineheight{1.25}\smash{\begin{tabular}[t]{l}$\mathtt{M}$\end{tabular}}}}%
  \end{picture}%
\endgroup%

%% file: content/diagnosis.tex
\section{DrPlanner}\label{sec:dr}
This section presents our prompt engineering with a nuanced diagnostic description. We begin by introducing the overall algorithm, followed by a more detailed presentation.


\renewcommand{\algorithmicrequire}{\textbf{Input:}}
\renewcommand{\algorithmicensure}{\textbf{Output:}}
\begin{figure}[!t]
	\vspace{-2mm}
\begin{algorithm}[H]
	\small
	\begin{algorithmic}[1]
		\Require planning problem $\mathtt{{P}}$, motion planner $\mathtt{M}$, target value $J^*$, system prompt $\ell_{\text{system}}$, LLM
		\Ensure diagnoses and prescriptions $\boldsymbol{\ell}_{\text{dp}}^*$, repaired planner $\mathtt{M}_{\text{rep}}^*$
		\State $\chi$ $\gets$ \Call{$\mathtt{M}$.plan}{$\mathtt{{P}}$}\label{alg1:plan}\vspace{.1mm}
		\State $J$ $\gets$ \Call{evaluate}{$\chi$}\label{alg1:eval}\vspace{.1mm}
		\State $\ell_{\text{user}}$ $\gets$ \Call{describe}{$\mathtt{M}$, $J$, $J^*$}\Comment{Sec. \ref{subsec:diag_des}} \label{alg1:describe} \vspace{.1mm}
		\State $J_{\min}$ $\gets$ $J$, $\boldsymbol{\ell}_{\text{dp}}^*$ $\gets$ $\emptyset$, $\mathtt{M}_{\text{rep}}^*$  $\gets$ $\emptyset$\vspace{.1mm}
		\While{not \Call{reachTokenLimit}{LLM} and $J_{\min}-J^*>\epsilon$ }\vspace{.1mm}\label{alg1:loop_start}
				\State $(\boldsymbol{\ell}_{\text{dp}}$, $\boldsymbol{p}_{\text{p}})$ $\gets$ LLM.\Call{query}{$\ell_{\text{system}}$, $\ell_{\text{user}}$} \Comment{Sec. \ref{subsec:reprompt}\enspace\!}\vspace{.01mm}\label{alg1:query_0}
				\State $\mathtt{M}_{\text{rep}}$ $\gets$ \Call{repair}{$\mathtt{M}$, $\boldsymbol{p}_{\text{p}}$} \label{alg1:repair_0}\vspace{.1mm} 
				\State $\chi$ $\gets$ $\mathtt{M}_{\text{rep}}$.\Call{plan}{$\mathtt{{P}}$}\label{alg1:plan_rep}\vspace{.1mm}
				\State $J_{\text{rep}}$ $\gets$ \Call{evaluate}{$\chi$}\label{alg1:eval_rep}\vspace{.1mm}
				\State $\ell_{\text{user}}$ $\gets$ \Call{addFeedback}{$\ell_{\text{user}}$, $J$, $J_{\text{rep}}$, $\boldsymbol{\ell}_{\text{dp}}$}\Comment{Sec. \ref{subsec:reprompt}} \vspace{.1mm}\label{alg1:add_feedback}
				\If{$J_{\text{rep}}<J_{\min}$}\vspace{.1mm}
						\State $J_{\min}$ $\gets$ $J_{\text{rep}}$, $\boldsymbol{\ell}_{\text{dp}}^*$ $\gets$ $\boldsymbol{\ell}_{\text{dp}}$, $\mathtt{M}_{\text{rep}}^*$$\gets$   $\mathtt{M}_{\text{rep}}$\vspace{.1mm}
				\EndIf\vspace{.1mm}
		\EndWhile\label{alg1:loop_end}
		\State \Return $\boldsymbol{\ell}_{\text{dp}}^*$, $\mathtt{M}_{\text{rep}}^*$ \vspace{-.5mm}\label{alg1:return}
	\end{algorithmic}
	\caption{\small\textsc{diagnoseAndRepairPlanner}}
	\label{alg:main}
\end{algorithm}
\vspace{-0.6cm}
\end{figure}
\subsection{Overall Algorithm}
A general overview of using $\mathtt{DrPlanner}$ 
is presented in Fig.~\ref{fig:framework} and Alg.~\ref{alg:main}. {Before initiating the process, the user fills in the placeholders enclosed in square brackets.}
For a given scenario, the motion planner $\mathtt{M}$ is first deployed to address the associated planning problem $\mathtt{P}$ (see line~\ref{alg1:plan}). Subsequently, the planned trajectory $\chi$ is evaluated using the objective function $J$ (see line~\ref{alg1:eval}). Following this, a diagnostic description $\ell_{\text{user}}$ encompassing the diagnostic instructions, the description of the planner, the evaluation of the trajectory, and the few-shot examples are formulated (see line~\ref{alg1:describe}). This description, along with the system prompt $\ell_{\text{system}}$, is fed into the LLM (see line~\ref{alg1:query_0}). The structure of the input prompt is illustrated in the center of the framework in Fig.~\ref{fig:framework}. Afterwards, the obtained patched programs are applied to the motion planner by integrating the modifications into the existing codebase (see line~\ref{alg1:repair_0}). 

However, it is important to note that the output generated may include errors such as hallucinations and inaccurate analyses \cite{yang2023harnessing}. 
To mitigate these issues, we employ an iterative prompting strategy, repeatedly refining the process. The iteration is terminated when a notable improvement in the planner is observed, e.g., when the difference between the current best performance $J_{\min}$ and a target value $J^*$ is smaller than a threshold $\epsilon\in\mathbb{R}_+$, or when the token limit of the LLM is reached (see lines~\ref{alg1:loop_start}-\ref{alg1:loop_end}). Finally, the repaired planner demonstrating the best improvement, if any, along with the corresponding diagnoses and prescriptions, is returned (see line~\ref{alg1:return}).

Another regime is to finetune the LLM to the given task. However, to date, there exists no open-source dataset containing input-output examples of motion planners. Additionally, finetuning usually only provides modest improvements in solving challenging and complex tasks compared to in-context learning \cite{wei2022chain, jain2023llm, chen2023teaching}. {Regardless of the approach, when deploying the repaired planners on roads, a safety layer is always required \cite{mehdipour2023formal}.}
\subsection{Diagnostic Description}\label{subsec:diag_des}
As discussed in Sec.~\ref{subsec:prompt_eng}, prompt design is challenging, particularly when considering the limited information about the diagnostic object in the pretrained LLM. To enhance conclusions, we design a structured and 
comprehensive description of the motion planner, emulating the process of a real doctor.  Its overall skeleton is depicted in the lower part of Fig.~\ref{fig:framework}.
As we assume that the motion planner internally handles goal-reaching and drivability-checking of the trajectory in the scenario  (cf. Sec. \ref{subsec:mp}), a detailed description of the scenario, motion planning problem, and trajectory states is omitted in the prompt. Alternatively, these tasks can be addressed by additional modules, such as those employing LLM-embedded agents (cf. Sec.~\ref{subsec:llm}).
\subsubsection{Instructions}\label{subsec:dia_instruction}
The instruction provides general guidance for the LLM, detailing the expected output and reasoning constraints. 
In addition, we can include the commonly used rule-of-thumb from expert knowledge. For instance, ``\textit{merely adjusting the weighting or coefficients is often cumbersome and not very effective}".
  
\subsubsection{Motion Planner}\label{subsec:dia_planner}
The description of the motion planner begins with the selection and a brief introduction to the planning algorithm. This is followed by a general description of the key components that primarily affect the performance of the planner. To gain a better understanding of how the algorithm is practically implemented, we also include the code of the key components as an additional input modality.
As mentioned in Sec. \ref{subsec:apr}, the LLM is then able to generate repaired programs given corresponding instructions.
{Motivated by the chain of thought (cf. Sec.~\ref{subsec:prompt_eng}), we incorporate existing explanations found within the docstrings of subfunctions to provide natural language summaries for the code blocks. The description adheres to the format of \textbf{\{subfunction name\}} followed by its \textbf{\{docstring\}}.} For instance, an automatically generated \textbf{\{detailed description\}} is: ``\texttt{\small \textcolor{black}{self}.calc\_angle\_to\_goal} \textit{returns the orientation of the goal with respect to current position; ...}" (cf. Fig.~\ref{fig:key_search}). 
\subsubsection{Planned Trajectory} \label{subsec:des_traj} There are various measures to quantitatively evaluate the planned trajectory and track its improvement. These measures include the cost function \cite{Althoff2017a}, criticality measures \cite{lin2023commonroad}, courtesy to other traffic participants \cite{schwarting2019social}, and degree of traffic rule compliance \cite{YuanfeiLin2022a, YuanfeiLinMPR}. To align the LLM with the desired behavior, we present not only the evaluation results for the selected measures but also incorporate the target value $J^*$, which can be, e.g., sourced from the motion planning benchmark leaderboard. In addition, the numerical data of the values and weights of the objective components is translated into a narrative description  by mapping them to their corresponding placeholders. 
\subsubsection{Few-Shots}\label{subsec:few-shots}
As it is not necessary for LLMs to have prior knowledge of the other part of the large-scale motion planner, we provide existing helper functions and their exemplary usage in the prompt. Furthermore, several human-annotated examples for improving the performance of the specific type of motion planner can be added here, with examples available in Fig.~\ref{fig:example}.




\newcommand{\mymintedfontsizes}{\fontsize{5.85pt}{8pt}\selectfont}
\subsection{LLM Querying and Iterative Prompting} \label{subsec:reprompt}
{When querying the LLM, it is essential to specify the desired output format. To achieve this, one can guide the LLM by emphasizing the diagnoses, prescriptions, and key components of the planner (cf. Sec.~\ref{subsec:mp}) in the prompt as desired responses  or employ other third-party tools such as LangChain\footnote{\footnotesize\url{https://www.langchain.com/}}. Consequently, the structured patched results can directly replace the original elements to repair the planner.}

Motivated by how LLMs are utilized in improving technical systems \cite{liu2023reflect, skreta2023errors, chen2023teaching, wen2023dilu}, we examine the repaired planner by executing it and then pass the evaluation result back to the LLM. In case of compilation or execution errors, the previous diagnostic result is combined with the information indicating where the error occurred and what it entails. Otherwise, the combination is made with a comparison of the performance between the updated planned trajectory and the original one. 

%% file: content/evaluation.tex
\section{Evaluation}\label{sec:eva}
We evaluate our approach using the open-source motion planners
from the CommonRoad platform \cite{Althoff2017a}, which are written in \texttt{Python}.  
As CommonRoad provides customizable challenges and annual competitions, where users can compete against each other on predefined benchmarks, we can continuously integrate enhancements into $\mathtt{DrPlanner}$ based on insights from a broad user base.  
Furthermore, we choose GPT-4-Turbo\footnote{ID \texttt{gpt-4-turbo-preview} in the API of OpenAI.} as our LLM and use its function calling feature to generate structured outputs. 
{It should be noted that our framework is not limited to GPT-4-Turbo and can be easily adapted for use with other LLMs by modifying the interface.}
The patched programs are then stringified in a JSON object and directly parsed to the motion planner, followed by execution through the \texttt{exec} function in \texttt{Python}. 
The token limit is set to $8,000$, the threshold $\epsilon$ is equal to $10$, and we choose the sampling temperature of the LLM at $0.6$ (cf. \cite[Fig.~5]{chen2021evaluating}). Code and exemplary prompts are available at {\url{https://github.com/CommonRoad/drplanner}}.


%
%
%
\subsection{Setup}
\subsubsection[Search-Based Motion Planner]{Search-Based Motion Planner\footnote{\footnotesize \url{https://commonroad.in.tum.de/tools/commonroad-search}}}
{We adapt the anytime A* search algorithm using lattice-based graphs \cite{pivtoraiko2011kinodynamic}. This implementation features a time-limited search cut-off and employs a cost function} and an estimated cost to the goal, namely, a \textit{heuristic function}, to guide the search process. The graph is constructed with \textit{motion primitives}—short trajectories generated offline through a forward simulation of a given vehicle model. {The number of explored nodes in the graph is denoted as $N_n$.}  Motion primitives are typically referenced by IDs encoded with configurable parameters\footnote{All parameters are given in SI units.}:
\begin{align*} \small
		\vspace{-2mm}
	\mathtt{MP} = \texttt{"\text{V}\_\colorbox{openaiblue!20}{\strut ${v_{\min}}$}\_\colorbox{openaiblue!20}{\strut $v_{\max}$}\_\text{Vstep}\_\colorbox{openaiblue!20}{\strut ${\Delta v}$}\_\text{SA}\_\colorbox{openaiblue!20}{\strut $\delta_{\min}$}\_\colorbox{openaiblue!20}{\strut $\delta_{\max}$}\_}\\[-1mm] \small
	\texttt{\text{SAstep}\_\colorbox{openaiblue!20}{\strut $\Delta \delta$}\_\text{T}\_\colorbox{openaiblue!20}{\strut $\tau$}\_\text{Model}\_\colorbox{openaiblue!20}{\strut $m$}"}\text{,}	\vspace{-2mm}
\end{align*}
where $v_{\min}$ and $v_{\max}$ are the sampling velocity limits, $\delta_{\min}$ and $\delta_{\max}$ are the sampling steering angle bounds, $\Delta v$ and $\Delta \delta$ specify their respective step sizes, $\tau$ is the time duration of each motion primitive, and $m$ is the model identifier of the ego vehicle. Therefore, the heuristic function and motion primitives constitute the key components. 
We provide the entire code block of the heuristic function along with descriptions of the involved subfunctions in natural language. 
In the description of motion primitives, the explanation includes the naming convention, followed by their ID. 


\subsubsection[Sampling-Based Motion Planner]{Sampling-Based Motion Planner\footnote{\footnotesize \url{https://commonroad.in.tum.de/tools/commonroad-reactive-planner}}}
{Similarly, we evaluate our approach on the sampling-based motion planner of~\cite{Werling2010}, which computes jerk-optimal trajectories using polynomials to connect sampled end states with the initial state. From the set of feasible trajectory samples, the optimal trajectory is selected based on a \textit{cost function}. Consequently, the cost function and \textit{sampling configurations}, such as the sampling time horizon $t_s$, are the key components.}

\subsubsection{Measures of the Planned Trajectory}
To evaluate the quality of the planned trajectory, we utilize the standardized objective function $J_{\text{SM1}}$\footnote{{The objective function can be adapted or replaced as needed.}} from CommonRoad~\cite[Sec. VI]{Althoff2017a}, which includes the cost for acceleration, steering angle, steering rate, distance and orientation offset to the centerline of the road, and velocity offset to the desired value. 
\renewcommand{\arraystretch}{1.02}
\newcolumntype{P}[1]{>{\centering\arraybackslash}p{#1}}
\subsubsection{Few-Shots}
To gain a deeper insight into the planner, we include method definitions and docstrings for existing helper functions within the planner class. 
For instance, as shown in Fig.~\ref{fig:example}, we also provide a list of IDs corresponding to offline-generated motion primitives from which the LLM can select for the search-based planner.

\usemintedstyle{vs}
\setminted{fontsize=\mymintedfontsizes} 
\setlength{\fboxsep}{0.5pt}
\begin{figure}[t!]
	\centering
	\vspace{0mm}
	\begin{minted}[mathescape, %bgcolor={openaigreen!20},
		numbersep=2pt,
		frame=lines,
		breaklines,
		framesep=01mm]{python3}
		|\texttt{There are some pre-defined helper functions that can be directly called}|
		|\texttt{in the \textbf{heuristic function}}|:
		def calc_acceleration_cost(self, path: List[KSState]) -> float:
		    """Returns the acceleration costs.""" ...
		|\texttt{\textbf{Examples}}|:
		(input) 
		def heuristic_function(self, node_current: PriorityNode) -> float:
		    ...
		    cost = angle_to_goal 
		    return cost
		(output)
		|\texttt{\textbf{Diagnosis}: the acceleration is not considered}|
		|\texttt{\textbf{Prescription}: add the acceleration cost to the heuristic function}|
		def heuristic_function(self, node_current: PriorityNode) -> float:
		    acceleration_cost = self.calc_acceleration_cost(node_current.list_paths[-1]) 
		    ...
		    cost = angle_to_goal + acceleration_cost
		    return cost
		|\texttt{Feasible \textbf{motion primitives} with the same name format that you can }| 
		|\texttt{directly use:}|
		"V_0.0_20.0_Vstep_1.0_SA_-1.066_1.066_SAstep_2.13_T_0.5_Model_BMW_320i",
		"V_0.0_20.0_Vstep_2.0_SA_-1.066_1.066_SAstep_0.18_T_0.5_Model_BMW_320i", 
		...
	\end{minted}
	\caption{Snippet of the few-shot prompting used for the search-based planner.}\label{fig:example}\vspace{-2mm}
\end{figure}

\begin{figure}[!t]%
	\centering
	\vspace{-0.5mm}
	\begin{subfigure}[!t]{\columnwidth}\centering
		\def\svgwidth{0.93\columnwidth}\footnotesize
		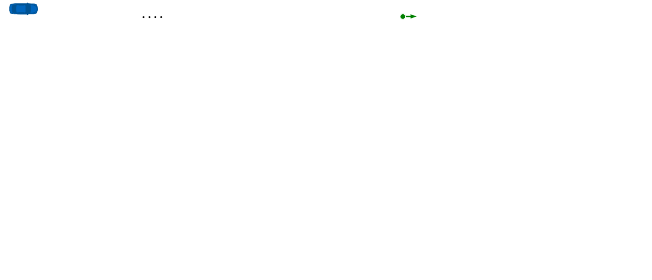
		\vspace{-1mm}
		\caption{Search-based motion planner.}\label{fig:search}
	\end{subfigure}
	\begin{subfigure}[!t]{\columnwidth}\centering
		\def\svgwidth{0.93\columnwidth}\footnotesize
		\hspace{-1mm}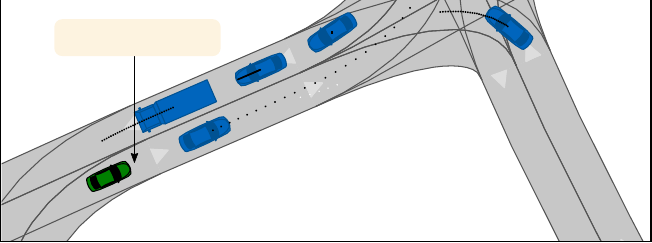
		\caption{{Sampling-based motion planner.}}\label{fig:sample}
		\vspace{-1mm}
	\end{subfigure}
	\caption{Critical intersection scenario\protect\footnotemark \ in which the ego vehicle needs to safely drive for $33$ time steps. For clarity, the planned trajectories for the ego vehicle from different planners are marked with different colors and labels.}\label{fig:repaired_planner}
	\vspace{-5mm}
\end{figure}%
\footnotetext{CommonRoad-ID: DEU\_Guetersloh-15\_2\_T-1}

\usemintedstyle{vs}
\newcommand{\mymintedfontsize}{\fontsize{5.85pt}{8pt}\selectfont}
\setlength{\fboxsep}{0.5pt}
\begin{figure}[t]
	\centering
	\vspace{0mm}
		\begin{subfigure}[!t]{\columnwidth}\centering
	\begin{minted}[mathescape,
		linenos,
		numbersep=2pt,
		frame=lines,
		breaklines,
		framesep=01mm]{python3}
def heuristic_function(self, node_current: PriorityNode) -> float:
    path_last = node_current.list_paths[-1]
    angleToGoal = self.calc_angle_to_goal(path_last[-1])		
    orientationToGoalDiff = self.calc_orientation_diff(angleToGoal, path_last[-1].orientation)
    cost_time = self.calc_time_cost(path_last)		
    if self.reached_goal(node_current.list_paths[-1]):
        heur_time = 0.0		
    if self.position_desired is None:
        heur_time = self.time_desired.start - node_current.list_paths[-1][-1].time_step		
    else:
        velocity = node_current.list_paths[-1][-1].velocity
    if np.isclose(velocity, 0):
        heur_time = np.inf	
    else:
        heur_time = self.calc_euclidean_distance(current_node=node_current) / velocity
    cost = 20 * orientationToGoalDiff + 0.5 * cost_time + heur_time
    if cost < 0:
        cost = 0
    return cost
	\end{minted}
\vspace{-\baselineskip} 
	\begin{minted}[mathescape,
			frame=lines,
			breaklines,
			framesep=1mm]{python3}
MP = "V_0.0_20.0_Vstep_4.0_SA_-1.066_1.066_SAstep_0.18_T_0.5_Model_\
    BMW_320i"
		\end{minted}
		\caption{Search-based motion planner.}\label{fig:key_search}
	\end{subfigure}
			\begin{subfigure}[!t]{\columnwidth}\centering
		\begin{minted}[mathescape,
			linenos,
			numbersep=2pt,
			frame=lines,
			breaklines,
			framesep=01mm]{python3}
		def evaluate(self, trajectory: TrajectorySample) -> float:
		    cost == 0.0
		    cost += CostFunction.steering_velocity_costs(trajectory)
		    cost += CostFunction.path_length_costs(trajectory)
		return cost
		\end{minted}
		\vspace{-\baselineskip} 
		\begin{minted}[mathescape,
			frame=lines,
			breaklines,
			framesep=1mm]{python3}
			planning:
			    dt: 0.1
			    replanning_frequency: 3
			    time_steps_computation: 20
		\end{minted}
		\caption{{Sampling-based motion planner.}}\label{fig:key_sample}
	\end{subfigure}
	\caption{{Key components} used in the initial planner.}\label{fig:heur_func_original}\vspace{-3.mm}
\end{figure}
\setlength{\fboxsep}{0.4pt}
\newcolumntype{Y}{>{\hsize=.44\hsize}X}
\newcolumntype{Z}{>{\hsize=.6\hsize}X}
\newcolumntype{H}{>{\hsize=.5	\hsize}X}
\subsection{Case Study}\label{subsec:case_study}

We choose an intersection scenario from the CommonRoad platform (cf.~Fig.~\ref{fig:repaired_planner}), which is generated by the scenario factory for safety-critical traffic scenarios \cite{klischat2020scenario, lin2023commonroad}. In the urban environment, the motion planners are responsible for navigating the ego vehicle from the initial state for $3.3s$ without colliding with any obstacles. The time increment of the scenario is $0.1s$. 
In both planned trajectories by the initial planners configured as shown in Fig.~\ref{fig:heur_func_original}, the ego vehicle brakes and steers slightly to the right, leading to high values of $J_{\text{SM1}}$ (cf. Tab.~\ref{tab:comparison_cost}). The target value of $J_{\text{SM1}}$ is extracted from the CommonRoad benchmark leaderboard\footnote{\footnotesize\url{https://commonroad.in.tum.de/solutions/ranking}} and is $J_{\text{SM1}}^*=0.16$. 

\begin{table}[t!]\footnotesize
	\begin{center}
		\vspace{0.5mm}
		\caption{Comparison of the planned trajectories before and after repair. The lowest values of the objective function are marked in bold.
		}
		\label{tab:comparison_cost}
		\begin{tabular}{p{1.1cm}P{0.5cm}>{\columncolor{initialplanner}}P{0.9cm}>{\columncolor{firstiteration}}P{0.8cm}>{\columncolor{seconditeration}}P{0.8cm}>{\columncolor{thirditeration}}P{0.8cm}>{\columncolor{fourceiteration}}P{0.8cm}}
			\toprule[1.pt]
			\rowcolor{white} \multirow{2}{*}{\cellcolor{white}\textbf{Type}} &  \multirow{2}{*}{\cellcolor{white}\textbf{Item}}  & \textbf{Initial} & \multicolumn{4}{c}{\textbf{Repaired Planner}} \\
			& & \cellcolor{white} \textbf{Planner} & {\cellcolor{white} \textbf{1. Iter.}} & {\cellcolor{white} \textbf{2. Iter.}}  & {\cellcolor{white} \textbf{3. Iter.}}  & {\cellcolor{white} \textbf{4. Iter.}} \\
			\midrule
			\multirow{2}{*}{\makecell[l]{\textbf{Search-}\\\textbf{based}}} 	
			& $J_\text{SM1}$ & $4606.93$ & { \cellcolor{firstiteration}$752.56$} & - & {\cellcolor{thirditeration}$\boldsymbol{4.65}$} & - \\
			& $N_n$ & $11$ & {\cellcolor{firstiteration}$9$} & - & {\cellcolor{thirditeration}$9$} & - \\
			\midrule
			\multirow{2}{*}{\makecell[l]{\textbf{{Sampling-}}\\\textbf{{based}}}}
			& $J_\text{SM1}$ & $2614.76$ & $169.49$ & $305.42$ &\!$\boldsymbol{147.55}$& $197.58$\\
			& $t_s$ & $2.0 s$ & $3.0 s$ & $2.5 s$ & $3.0 s$ & $3.0 s$\\
			\bottomrule
		\end{tabular}
		\vspace{-7mm}
	\end{center}
\end{table}

The diagnostic results {for the search-based planner} using our approach are illustrated in Fig.~\ref{fig:diag_results}. In the first iteration, the provided helper functions are automatically included in the heuristic function by the LLM (cf. Fig.~\ref{fig:1iteration}). Meanwhile, some hyperparameters are adjusted, such as the orientation weight and the heuristic for zero velocity, and new motion primitives are selected. 
Considering all the above factors, the repaired planner results in a decrease in $J_{\text{SM1}}$ of the planned trajectory, particularly in the acceleration cost. {Additionally, it allows the vehicle to travel further forward with fewer explored nodes in the search graph due to the coarser motion primitives applied (see Tab.~\ref{tab:comparison_cost} and Fig.~\ref{fig:search}).}
In contrast, the diagnostic result from the second iteration leads to a \texttt{KeyError} (cf. Fig.~\ref{fig:2iteration}), indicating that the repaired heuristic function is not provided by the LLM.
With the iterative prompting, the error message is incorporated as feedback into the prompt for the third iteration. As shown in Fig.~\ref{fig:3iteration}, our approach not only helps the LLM avoid the errors from previous iterations (cf. the diagnosis ``\textit{KeyError in heuristic function}") but also retains the previous diagnostic results that lead to a positive impact on the planner. As a result, the planner significantly improves its performance, with a substantial reduction in $J_{\text{SM1}}$ from $752.56$ to $4.65$, achieved by further balancing the objective components (cf. Tab.~\ref{tab:comparison_cost}).
Moreover, it can be observed from Fig.~\ref{fig:diag_results} that $\mathtt{DrPlanner}$ can provide fine-grained diagnoses and prescriptions based on both the prompt design and fundamental aspects of programming, such as aliasing (cf. lines 10, 13, 15 in Fig.~\ref{fig:3iteration}). The resulting patched programs align precisely with these diagnoses and prescriptions. 

{The initial configuration snippet of the sampling-based planner is shown in Fig.~\ref{fig:key_sample}. A similar repair pattern to the search-based planner can be observed in Tab.~\ref{tab:comparison_cost} and Fig.~\ref{fig:sample}.  For brevity, we only show the diagnostic details for the third iteration in  Fig.~\ref{fig:diag_results_sample}, which achieves the best performance among all iterations. The cost function improves through weight tuning and adding more items, and a larger $t_s$ is selected, leading to a noticeable reduction in $J_{\text{SM1}}$.} 
\subsection{Performance Evaluation}
We further evaluate the performance of $\mathtt{DrPlanner}$ {by analyzing $50$ randomly selected critical CommonRoad scenarios, 
along with $50$ A*-search-based motion planners in various setups from the CommonRoad challenges. {The former is benchmarked against the search-based planner configured as shown in Fig.~\ref{fig:key_search}. The latter evaluation utilizes the scenario illustrated in Fig.~\ref{fig:repaired_planner}} and employs the pass@$k$ metric. We use its unbiased version as proposed in \cite[Sec.~2.1]{chen2021evaluating}, defined as the probability that at least one of the top $k \in \mathbb{N}_+$ generated code samples for a problem passes the given tests.
Here, we use a decrease of $J_{\text{SM1}}$ for the returned planner as the criterion for passing. {As a baseline, the performance of $\mathtt{DrPlanner}$ is compared with a genetic approach \cite{weimer2009automatically}, where the program of the heuristic function is repaired to minimize $J_{\text{SM1}}$. The solution space consists of $10$ chromosomes, and the process runs for $100$ generations.}
Additionally, we conduct ablation studies to examine the impact of omitting two specific components within the framework {across different planners}: few-shots and feedback. For each study, we execute the framework $10$ times to collect solution samples. 

\begin{figure}\centering
	    \begin{minipage}[t!]{\columnwidth}
\begin{subfigure}[t!]{\columnwidth}	\centering
	\centering
	\begin{minted}[mathescape,
		linenos,
		numbersep=2pt,
		frame=lines,
		breaklines,
		framesep=1mm]{python3}
		def heuristic_function(self, node_current: PriorityNode) -> float:
		|\colorbox{firstiteration!}{\strut+   acceleration\_cost = self.calc\_acceleration\_cost(path\_last)}||\setcounter{FancyVerbLine}{6}|
		|\colorbox{firstiteration!}{\strut+   path\_efficiency = self.calc\_path\_efficiency(path\_last)}|
		|\colorbox{firstiteration!}{\strut+   steering\_angle\_cost = self.calc\_steering\_angle\_cost(path\_last)}|
		|\colorbox{firstiteration!}{\strut+   steering\_velocity\_cost = self.calc\_steering\_velocity\_cost(path\_last)}|
		    if self.reached_goal(node_current.list_paths[-1]):
		|\colorbox{openaigreen!50}{\strut-       heur\_time = 0.0}|
		|\colorbox{firstiteration!}{\strut+       return 0}||\setcounter{FancyVerbLine}{11}|
		    if np.isclose(velocity, 0):|\setcounter{FancyVerbLine}{16}|    
		|\colorbox{openaigreen!50}{\strut-       heur\_time = np.inf}|
		|\colorbox{firstiteration!}{\strut+       heur\_time = 10}||\setcounter{FancyVerbLine}{17}|  # Instead of np.inf, assign a high but finite cost	
		|\colorbox{openaigreen!50}{\strut-   cost = 20 * orientationToGoalDiff + 0.5 * cost\_time + heur\_time}||\setcounter{FancyVerbLine}{20}|
		|\colorbox{firstiteration!}{\strut+   \parbox[t]{6.55cm}{cost = 10 * orientationToGoalDiff + 0.5 * cost\_time + \\ \parbox[t]{0.35cm}{ } heur\_time + acceleration\_cost + path\_efficiency +\\ \parbox[t]{0.35cm}{ } steering\_angle\_cost + steering\_velocity\_cost}}||\setcounter{FancyVerbLine}{20}|	
	\end{minted}
\vspace{-\baselineskip} 
\begin{minted}[mathescape,
	frame=lines,
	breaklines,
	framesep=1mm,
	escapeinside=@@]{python3}
	MP = @\textquotedblright@@\texttt{V\_0.0\_20.0\_Vstep\_}@@\colorbox{openaigreen!50}{\strut\st{4.0}}\colorbox{firstiteration}{\strut2.0}@@\texttt{\_SA\_-1.066\_1.066\_SAstep\_0.18\_T\_0.5\_Model\_}@\
	    BMW_320i@\textquotedblright@
\end{minted}
\vspace{-0.2\baselineskip} 
\scriptsize{
	{\renewcommand{\arraystretch}{1.15} 
\begin{tabularx}{\columnwidth} {YZ}

	\textbf{Diagnosis} & \textbf{Prescription} \\
	\hline
	Orientation weight too high  &  Adjust the weight for orientation to goal difference \\
	Missing acceleration cost &  Include acceleration cost in heuristic \\
	Missing path efficiency &  Include path efficiency in heuristic \\
	Missing steering angle cost &  Include steering angle cost in heuristic \\
	Missing steering velocity cost &  Include steering velocity cost in heuristic \\
	Infinite heuristic for zero velocity &  Handle zero velocity case appropriately \\
	Sparse motion primitives & Recommend motion primitives with higher branching factor \\
	\hline
\end{tabularx}
}}
\vspace{-0.3\baselineskip} 
\caption{1. iteration.}\label{fig:1iteration}
\end{subfigure}
\begin{subfigure}[!t]{\columnwidth}
\begin{minted}[mathescape,
	frame=lines,
	breaklines,
	framesep=1mm,
	escapeinside=@@]{python3}
	KeyError: 'repaired_heuristic_function'
\end{minted}
	\vspace{-0.3\baselineskip} 
	\caption{2. iteration.}\label{fig:2iteration}
\end{subfigure}

\begin{subfigure}[!t]{\columnwidth}	\centering
	\begin{minted}[mathescape,
		linenos,
		numbersep=2pt,
		frame=lines,
		breaklines,
		framesep=1mm]{python3}
		def heuristic_function(self, node_current: PriorityNode) -> float:
		|\fcolorbox{black}{thirditeration!60}{\strut+   acceleration\_cost = self.calc\_acceleration\_cost(path\_last)}||\setcounter{FancyVerbLine}{6}|
		|\fcolorbox{black}{thirditeration!60}{\strut+   path\_efficiency = self.calc\_path\_efficiency(path\_last)}|
		|\fcolorbox{black}{thirditeration!60}{\strut+   steering\_angle\_cost = self.calc\_steering\_angle\_cost(path\_last)}|
		|\fcolorbox{black}{thirditeration!60}{\strut+   steering\_velocity\_cost = self.calc\_steering\_velocity\_cost(path\_last)}|
		|\colorbox{openaigreen!50}{\strut-   if self.reached\_goal(node\_current.list\_paths[-1]):}|
		|\colorbox{thirditeration!60}{\strut+   if self.reached\_goal(path\_last):}||\setcounter{FancyVerbLine}{10}|
		        heur_time = 0.0
		|\colorbox{openaigreen!50}{\strut-   if self.position\_desired is None:}|
		|\colorbox{thirditeration!60}{\strut+   elif self.position\_desired is None:}||\setcounter{FancyVerbLine}{12}|
		|\colorbox{openaigreen!50}{\strut-       \parbox[t]{5.6cm}{heur\_time = self.time\_desired.start -  \\ \parbox[t]{0.35cm}{ } node\_current.list\_paths[-1][-1].time\_step}}|
		|\colorbox{thirditeration!60}{\strut+       heur\_time = self.time\_desired.start - path\_last[-1].time\_step}||\setcounter{FancyVerbLine}{13}|
		    else:
		|\colorbox{openaigreen!50}{\strut-       velocity = node\_current.list\_paths[-1][-1].velocity}|
		|\colorbox{thirditeration!60}{\strut+       velocity = path\_last[-1].velocity}| |\setcounter{FancyVerbLine}{15}|
		    if np.isclose(velocity, 0):
		|\colorbox{openaigreen!50}{\strut-       heur\_time = np.inf}|
		|\colorbox{thirditeration!60}{\strut+       heur\_time = 1e6}||\setcounter{FancyVerbLine}{17}|  # A large but not infinite cost	
		|\fcolorbox{black}{openaigreen!50}{\strut-   cost = 20 * orientationToGoalDiff + 0.5 * cost\_time + heur\_time}||\setcounter{FancyVerbLine}{20}|
		|\fcolorbox{black}{thirditeration!60}{\strut+   \parbox[t]{6.55cm}{cost = 10 * orientationToGoalDiff + 0.5 * cost\_time + \\ \parbox[t]{0.35cm}{ } heur\_time + acceleration\_cost + path\_efficiency +\\ \parbox[t]{0.35cm}{ } steering\_angle\_cost + steering\_velocity\_cost}}||\setcounter{FancyVerbLine}{20}|	
		\end{minted}
\vspace{-\baselineskip} 
\begin{minted}[mathescape,
	frame=lines,
	breaklines,
	framesep=1mm,
	escapeinside=@@]{python3}
	MP = @\textquotedblright@@\texttt{V\_0.0\_20.0\_Vstep\_}@@\fcolorbox{black}{openaigreen!50}{\strut\st{4.0}}\fcolorbox{black}{thirditeration!60}{\strut2.0}@@\texttt{\_SA\_-1.066\_1.066\_SAstep\_0.18\_T\_0.5\_Model}@\
	    _BMW_320i@\textquotedblright@
\end{minted}
\vspace{-0.2\baselineskip} 
\scriptsize{
{\renewcommand{\arraystretch}{1.15} 
	\begin{tabularx}{\columnwidth} {HZ}
		
		\textbf{Diagnosis} & \textbf{Prescription} \\
		\hline
	Orientation weight excessive  &   Decrease orientation weight in heuristic \\
	Heuristic excludes costs  &  Incorporate all costs into heuristic  \\
	Zero velocity infinite cost &  Refine zero velocity case handling \\
	Motion primitives high branching  &  Optimize motion primitives branching \\
	KeyError in heuristic function &  Ensure correct key for improved heuristic\\
			\hline
	\end{tabularx}
}}
\vspace{-0.3\baselineskip} 
\caption{3. iteration.}\label{fig:3iteration}
\end{subfigure} 
\caption{Diagnostic and repair result for the search-based motion planner in Fig.~\ref{fig:key_search}. The identical program patches in the first and third iteration are highlighted with black borders in (c). For the second iteration, we omit the diagnoses and prescriptions since it leads to an error.}\label{fig:diag_results}
    \end{minipage}
    
    \begin{minipage}[t]{\columnwidth}
	\centering
	\vspace{3mm}
	\def\svgwidth{0.98\columnwidth}\scriptsize
	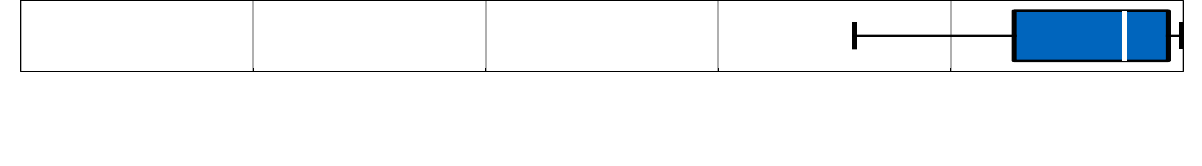
	\caption{{Benchmarked reduction of $J_{\text{SM1}}$ across scenarios using $\mathtt{DrPlanner}$. For better visibility, outliers in the box plot are not shown.}}\label{fig:more_scenarios}
	    \end{minipage}
\end{figure}%

\begin{figure}
		\vspace{0mm}
			\centering
			\begin{minted}[mathescape,
				linenos,
				numbersep=2pt,
				frame=lines,
				breaklines,
				framesep=1mm]{python3}
				def evaluate(self, trajectory: TrajectorySample) -> float:
				|\colorbox{openaigreen!50}{\strut-   cost += CostFunction.steering\_velocity\_costs(trajectory)}| |\setcounter{FancyVerbLine}{3}|
				|\colorbox{thirditeration!80}{\strut+   cost += CostFunction.acceleration\_costs(trajectory) * 25}||\setcounter{FancyVerbLine}{3}| 
				|\colorbox{thirditeration!80}{\strut+   cost += CostFunction.steering\_velocity\_costs(trajectory) * 25}|
				|\colorbox{thirditeration!80}{\strut+   cost += CostFunction.longitudinal\_jerk\_costs(trajectory) * 10}|
			\end{minted}
			\vspace{-\baselineskip} 
			\begin{minted}[mathescape,
				frame=lines,
				breaklines,
				framesep=1mm,
				escapeinside=@@]{python3}
				planning:
				    time_steps_computation: @\colorbox{openaigreen!50}{\strut\st{20}}\colorbox{thirditeration!80}{\strut30}@
			\end{minted}
			\vspace{-0.2\baselineskip} 
			\scriptsize{
				{\renewcommand{\arraystretch}{1.1} 
					\begin{tabularx}{\columnwidth} {HZ}
						
						\textbf{Diagnosis} & \textbf{Prescription} \\
						\hline
						Acceleration cost weight too high  &  Reduce the weight of the acceleration cost \\
						Steering velocity cost weight too high &  Consider reducing its weight to better balance the cost function \\
						Inclusion of jerk costs &  To add nuance to the trajectory evaluation, include jerk costs in the cost function \\
						Planning horizon too short &  Increase the planning horizon \\
						\hline
					\end{tabularx}
			}}
			\vspace{-0.3\baselineskip} 
		\caption{{Diagnostic and repair result of the third iteration for the sampling-based motion planner in Fig.~\ref{fig:key_sample}.}}\label{fig:diag_results_sample}\vspace{-2mm}
\end{figure}%
  {Fig.~\ref{fig:more_scenarios}} and Tab.~\ref{tab:performance} present the results of the performance evaluation. Overall, $\mathtt{DrPlanner}$ effectively diagnoses and repairs motion planners under various setups, {outperforming the baseline approach in all metrics, with a pass rate of $98.0\%$ at $k=10$ and an average reduction of $54.5\%$ in $J_{\text{SM1}}$.} {The benchmark results in Fig.~\ref{fig:more_scenarios} further indicate robust performance across diverse scenarios, with an average $J_{\text{SM1}}$ decrease of $90.76\%$.} 
  Note that, similar to the case study in Sec.~\ref{subsec:case_study}, the value of $J_{\text{SM1}}$ does not converge with the iterations due to diagnostic inaccuracies.  However, the average number of iterations required to observe its first decrease is $1.4$.
  
Moreover, the ablation studies demonstrate that both the few-shot learning (cf. Sec.~\ref{subsec:few-shots}) and the iterative prompting (cf. Sec.~\ref{subsec:reprompt}) play crucial roles in enhancing the effectiveness of $\mathtt{DrPlanner}$.  In particular, the few-shots prompting is more effective since the LLM is intrinsically unaware of the other supportive components of the planner, e.g., the available motion primitives. Additionally, since the initial planners are not buggy but underperforming, the results without using few-shots show that they cannot be easily improved with only the descriptions of the planner and the planned trajectory. {Likewise, this applies to the baseline, which only optimizes the existing code that already works.}


%

\renewcommand{\arraystretch}{1.}
\begin{table}[t!]\footnotesize
	\begin{center}
		\vspace{0.5mm}
		\caption{{Performance evaluation} and ablation studies {across planners} on the design of $\mathtt{DrPlanner}$. Values in bold denote the best performance.}
		\label{tab:performance}
		\begin{tabular}{p{1.7cm}P{0.79cm}P{0.9cm}P{0.9cm}P{0.6cm}P{1.cm}}
			\toprule[1.pt]
			\multirow{2}{*}{\textbf{Method}} & \multicolumn{3}{c}{\textbf{pass@$\boldsymbol{k}$} } &\multicolumn{2}{c}{\!\!\textbf{Decrement of $\boldsymbol{J_{\text{SM1}}}$}} \\
			& $\!k\!=\!1\!\uparrow\!\!$ & $\!k\!=\!5\!\uparrow\!$ & $\!\!k\!=\!10\!\uparrow\!\!$ &\!\!\!\! Avg. $\!\uparrow\!\!\!$ &\!\!\!\!\!\! Std. Dev.\\
			\midrule
			{Genetic \cite{weimer2009automatically}} \!\! & $0.8\%$ &  $0.4\%$ &  $7.7\%$ & $0.1\%$ & $1.6\%$ \\
			\midrule
			w/o Few-Shots\!\! & $0.0\%$  &$0.0\%$ &$0.0\%$ & $0.0\%$ & ${0.0\%}$\\
			w/o Feedback & $45.4\%$ & $86.2\%$ & $\!\!{92.0\%}$ & $\!49.6\%$ & ${36.3\%}$ \\
			\midrule
			$\mathtt{DrPlanner}$ & $\boldsymbol{68.0\%}$ & $\boldsymbol{95.1\%}$ & $\!\!\boldsymbol{98.0\%}$ & $\!\!\boldsymbol{54.5\%}$ & ${34.9\%}$ \\
			\bottomrule
		\end{tabular}
		\vspace{-7mm}
	\end{center}
\end{table}

%% file: figures/repaired_planner.pdf_tex
\begingroup%
  \makeatletter%
  \providecommand\color[2][]{%
    \errmessage{(Inkscape) Color is used for the text in Inkscape, but the package 'color.sty' is not loaded}%
    \renewcommand\color[2][]{}%
  }%
  \providecommand\transparent[1]{%
    \errmessage{(Inkscape) Transparency is used (non-zero) for the text in Inkscape, but the package 'transparent.sty' is not loaded}%
    \renewcommand\transparent[1]{}%
  }%
  \providecommand\rotatebox[2]{#2}%
  \newcommand*\fsize{\dimexpr\f@size pt\relax}%
  \newcommand*\lineheight[1]{\fontsize{\fsize}{#1\fsize}\selectfont}%
  \ifx\svgwidth\undefined%
    \setlength{\unitlength}{312.91577148bp}%
    \ifx\svgscale\undefined%
      \relax%
    \else%
      \setlength{\unitlength}{\unitlength * \real{\svgscale}}%
    \fi%
  \else%
    \setlength{\unitlength}{\svgwidth}%
  \fi%
  \global\let\svgwidth\undefined%
  \global\let\svgscale\undefined%
  \makeatother%
  \begin{picture}(1,0.42888948)%
    \lineheight{1}%
    \setlength\tabcolsep{0pt}%
    \put(0,0){\includegraphics[width=\unitlength,page=1]{repaired_planner.pdf}}%
    \put(0.076941,0.39404537){\color[rgb]{0,0,0}\makebox(0,0)[lt]{\lineheight{1.25}\smash{\begin{tabular}[t]{l}obstacles\end{tabular}}}}%
    \put(0.25925714,0.41051637){\color[rgb]{0,0,0}\makebox(0,0)[lt]{\lineheight{1.25}\smash{\begin{tabular}[t]{l}future\end{tabular}}}}%
    \put(0.25912626,0.37980826){\color[rgb]{0,0,0}\makebox(0,0)[lt]{\lineheight{1.25}\smash{\begin{tabular}[t]{l}movement\end{tabular}}}}%
    \put(0.50385737,0.41057547){\color[rgb]{0,0,0}\makebox(0,0)[lt]{\lineheight{1.25}\smash{\begin{tabular}[t]{l}ego\end{tabular}}}}%
    \put(0.50372649,0.37986736){\color[rgb]{0,0,0}\makebox(0,0)[lt]{\lineheight{1.25}\smash{\begin{tabular}[t]{l}vehicle\end{tabular}}}}%
    \put(0.64597405,0.41063121){\color[rgb]{0,0,0}\makebox(0,0)[lt]{\lineheight{1.25}\smash{\begin{tabular}[t]{l}initial\end{tabular}}}}%
    \put(0.64584317,0.37992309){\color[rgb]{0,0,0}\makebox(0,0)[lt]{\lineheight{1.25}\smash{\begin{tabular}[t]{l}state\end{tabular}}}}%
    \put(0.84110125,0.41053559){\color[rgb]{0,0,0}\makebox(0,0)[lt]{\lineheight{1.25}\smash{\begin{tabular}[t]{l}planned\end{tabular}}}}%
    \put(0.84097037,0.37982752){\color[rgb]{0,0,0}\makebox(0,0)[lt]{\lineheight{1.25}\smash{\begin{tabular}[t]{l}trajectories\end{tabular}}}}%
    \put(0,0){\includegraphics[width=\unitlength,page=2]{repaired_planner.pdf}}%
    \put(0.11183482,0.30541227){\color[rgb]{0,0,0}\makebox(0,0)[lt]{\lineheight{1.25}\smash{\begin{tabular}[t]{l}initial planner\end{tabular}}}}%
    \put(0,0){\includegraphics[width=\unitlength,page=3]{repaired_planner.pdf}}%
    \put(0.32744001,0.0540035){\color[rgb]{0,0,0}\makebox(0,0)[lt]{\lineheight{1.25}\smash{\begin{tabular}[t]{l}1. iter.\end{tabular}}}}%
    \put(0,0){\includegraphics[width=\unitlength,page=4]{repaired_planner.pdf}}%
    \put(0.5991121,0.05456633){\color[rgb]{0,0,0}\makebox(0,0)[lt]{\lineheight{1.25}\smash{\begin{tabular}[t]{l}3. iter.\end{tabular}}}}%
    \put(0,0){\includegraphics[width=\unitlength,page=5]{repaired_planner.pdf}}%
  \end{picture}%
\endgroup%

%% file: figures/repaired_planner_sample2.pdf_tex
\begingroup%
  \makeatletter%
  \providecommand\color[2][]{%
    \errmessage{(Inkscape) Color is used for the text in Inkscape, but the package 'color.sty' is not loaded}%
    \renewcommand\color[2][]{}%
  }%
  \providecommand\transparent[1]{%
    \errmessage{(Inkscape) Transparency is used (non-zero) for the text in Inkscape, but the package 'transparent.sty' is not loaded}%
    \renewcommand\transparent[1]{}%
  }%
  \providecommand\rotatebox[2]{#2}%
  \newcommand*\fsize{\dimexpr\f@size pt\relax}%
  \newcommand*\lineheight[1]{\fontsize{\fsize}{#1\fsize}\selectfont}%
  \ifx\svgwidth\undefined%
    \setlength{\unitlength}{312.95937606bp}%
    \ifx\svgscale\undefined%
      \relax%
    \else%
      \setlength{\unitlength}{\unitlength * \real{\svgscale}}%
    \fi%
  \else%
    \setlength{\unitlength}{\svgwidth}%
  \fi%
  \global\let\svgwidth\undefined%
  \global\let\svgscale\undefined%
  \makeatother%
  \begin{picture}(1,0.37130087)%
    \lineheight{1}%
    \setlength\tabcolsep{0pt}%
    \put(0,0){\includegraphics[width=\unitlength,page=1]{repaired_planner_sample2.pdf}}%
    \put(0.11414943,0.30384814){\color[rgb]{0,0,0}\makebox(0,0)[lt]{\lineheight{1.25}\smash{\begin{tabular}[t]{l}initial planner\end{tabular}}}}%
    \put(0,0){\includegraphics[width=\unitlength,page=2]{repaired_planner_sample2.pdf}}%
    \put(0.84789693,0.14204447){\color[rgb]{0,0,0}\makebox(0,0)[lt]{\lineheight{1.25}\smash{\begin{tabular}[t]{l}3. iter.\end{tabular}}}}%
    \put(0,0){\includegraphics[width=\unitlength,page=3]{repaired_planner_sample2.pdf}}%
    \put(0.4505612,0.07309508){\color[rgb]{0,0,0}\makebox(0,0)[lt]{\lineheight{1.25}\smash{\begin{tabular}[t]{l}1. iter.\end{tabular}}}}%
    \put(0,0){\includegraphics[width=\unitlength,page=4]{repaired_planner_sample2.pdf}}%
    \put(0.84641494,0.07574216){\color[rgb]{0,0,0}\makebox(0,0)[lt]{\lineheight{1.25}\smash{\begin{tabular}[t]{l}2. iter.\end{tabular}}}}%
    \put(0,0){\includegraphics[width=\unitlength,page=5]{repaired_planner_sample2.pdf}}%
    \put(0.45125334,0.14204115){\color[rgb]{0,0,0}\makebox(0,0)[lt]{\lineheight{1.25}\smash{\begin{tabular}[t]{l}4. iter.\end{tabular}}}}%
  \end{picture}%
\endgroup%

%% file: figures/benchmark.pdf_tex
\begingroup%
  \makeatletter%
  \providecommand\color[2][]{%
    \errmessage{(Inkscape) Color is used for the text in Inkscape, but the package 'color.sty' is not loaded}%
    \renewcommand\color[2][]{}%
  }%
  \providecommand\transparent[1]{%
    \errmessage{(Inkscape) Transparency is used (non-zero) for the text in Inkscape, but the package 'transparent.sty' is not loaded}%
    \renewcommand\transparent[1]{}%
  }%
  \providecommand\rotatebox[2]{#2}%
  \newcommand*\fsize{\dimexpr\f@size pt\relax}%
  \newcommand*\lineheight[1]{\fontsize{\fsize}{#1\fsize}\selectfont}%
  \ifx\svgwidth\undefined%
    \setlength{\unitlength}{576.8211545bp}%
    \ifx\svgscale\undefined%
      \relax%
    \else%
      \setlength{\unitlength}{\unitlength * \real{\svgscale}}%
    \fi%
  \else%
    \setlength{\unitlength}{\svgwidth}%
  \fi%
  \global\let\svgwidth\undefined%
  \global\let\svgscale\undefined%
  \makeatother%
  \begin{picture}(1,0.13762262)%
    \lineheight{1}%
    \setlength\tabcolsep{0pt}%
    \put(0,0){\includegraphics[width=\unitlength,page=1]{benchmark.pdf}}%
    \put(0.38485662,0.00399639){\color[rgb]{0,0,0}\makebox(0,0)[lt]{\lineheight{1.25}\smash{\begin{tabular}[t]{l}Decrement of ${J_{\text{SM1}}}$ \end{tabular}}}}%
    \put(-0.00048266,0.0422739){\color[rgb]{0,0,0}\makebox(0,0)[lt]{\lineheight{1.25}\smash{\begin{tabular}[t]{l}$\SI{0}{\!\percent}$\end{tabular}}}}%
    \put(0.18021578,0.04227562){\color[rgb]{0,0,0}\makebox(0,0)[lt]{\lineheight{1.25}\smash{\begin{tabular}[t]{l}$\SI{20}{\!\percent}$\end{tabular}}}}%
    \put(0.37375219,0.0421336){\color[rgb]{0,0,0}\makebox(0,0)[lt]{\lineheight{1.25}\smash{\begin{tabular}[t]{l}$\SI{40}{\!\percent}$\end{tabular}}}}%
    \put(0.56756187,0.04220131){\color[rgb]{0,0,0}\makebox(0,0)[lt]{\lineheight{1.25}\smash{\begin{tabular}[t]{l}$\SI{60}{\!\percent}$\end{tabular}}}}%
    \put(0.76169766,0.04226362){\color[rgb]{0,0,0}\makebox(0,0)[lt]{\lineheight{1.25}\smash{\begin{tabular}[t]{l}$\SI{80}{\!\percent}$\end{tabular}}}}%
    \put(0.92787955,0.04324871){\color[rgb]{0,0,0}\makebox(0,0)[lt]{\lineheight{1.25}\smash{\begin{tabular}[t]{l}$\SI{100}{\!\percent}$\end{tabular}}}}%
  \end{picture}%
\endgroup%

%% file: content/conclusion.tex
\vspace{-2mm}
\section{Conclusion}\label{sec:conc}
We present the first framework for diagnosing and repairing motion planners {for automated vehicles} that leverages both common sense and domain-specific knowledge about causal mechanisms in LLMs. Through a modular and iterative prompt design, our approach automates the generation of descriptions for the planner and continuously enhances diagnostic performance.
 The major limitation of our approach is that the improvement of the planner cannot be guaranteed.
 However, as the capabilities of LLMs advance, we anticipate the paradigm to enhance significantly over time.
{Future work will {involve conducting additional tests across various application domains} and developing datasets by monitoring user submissions over time}. 
{Additionally, we plan to extend the few-shot component with a memory module to leverage experiential learning.} 
 We encourage researchers using $\mathtt{DrPlanner}$ to refine their motion planners and contribute towards establishing a large-scale framework that encompasses a variety of planner types for diagnostic and repair tasks.
 